\documentclass[nohyperref]{article}

\usepackage[colorlinks=true, citecolor=teal]{hyperref}  
\usepackage{microtype}
\usepackage{graphicx}
\usepackage{subfigure}
\usepackage{url}
\usepackage{bm}
\usepackage{float}
\usepackage{caption}
\usepackage{algorithmic}
\usepackage{multirow}
\usepackage{fancyhdr}
\usepackage{booktabs}
\usepackage{enumitem}
\usepackage{amsmath}
\usepackage{amssymb}
\usepackage{mathtools}
\usepackage{amsthm}
\usepackage[table]{xcolor} 
\usepackage[textsize=tiny]{todonotes}
\usepackage[capitalize,noabbrev]{cleveref}
\usepackage{colortbl}
\usepackage[inkscapeformat=png]{svg}

\definecolor{LightGreen}{rgb}{0.93,0.98,0.96}

\usepackage[accepted]{icml2024}

\usepackage[capitalize,noabbrev]{cleveref}

\theoremstyle{plain}

\theoremstyle{definition}

\theoremstyle{remark}

\icmltitlerunning{Layer-Aware Analysis of Catastrophic Overfitting: Revealing the Pseudo-Robust Shortcut Dependency}

\begin{document}

\twocolumn[
\icmltitle{Layer-Aware Analysis of Catastrophic Overfitting:\\ Revealing the Pseudo-Robust Shortcut Dependency}


\begin{icmlauthorlist}
\icmlauthor{Runqi Lin}{usyd}
\icmlauthor{Chaojian Yu}{usyd}
\icmlauthor{Bo Han}{hkbu}
\icmlauthor{Hang Su}{tsinghua}
\icmlauthor{Tongliang Liu}{usyd}

\end{icmlauthorlist}
\icmlaffiliation{usyd}{Sydeny AI Centre, School of Computer Science, The University of Sydney, Sydney, Australia}
\icmlaffiliation{hkbu}{Department of Computer Science, Hong Kong Baptist University, Hong Kong, China}
\icmlaffiliation{tsinghua}{Department of Computer Science and Technology, Institute for AI,
 BNRist Center, Tsinghua University, Beijing, China}
\icmlcorrespondingauthor{Tongliang Liu}{tongliang.liu@sydney.edu.au}
\icmlkeywords{Machine Learning, ICML}
\vskip 0.3in
]



\printAffiliationsAndNotice{}  

\begin{abstract}
Catastrophic overfitting (CO) presents a significant challenge in single-step adversarial training (AT), manifesting as highly distorted deep neural networks (DNNs) that are vulnerable to multi-step adversarial attacks. However, the underlying factors that lead to the distortion of decision boundaries remain unclear. In this work, we delve into the specific changes within different DNN layers and discover that during CO, the former layers are more susceptible, experiencing earlier and greater distortion, while the latter layers show relative insensitivity. Our analysis further reveals that this increased sensitivity in former layers stems from the formation of \emph{pseudo-robust shortcuts}, which alone can impeccably defend against single-step adversarial attacks but bypass genuine-robust learning, resulting in distorted decision boundaries. Eliminating these shortcuts can partially restore robustness in DNNs from the CO state, thereby verifying that dependence on them triggers the occurrence of CO. This understanding motivates us to implement adaptive weight perturbations across different layers to hinder the generation of \emph{pseudo-robust shortcuts}, consequently mitigating CO. Extensive experiments demonstrate that our proposed method, \textbf{L}ayer-\textbf{A}ware Adversarial Weight \textbf{P}erturbation (LAP), can effectively prevent CO and further enhance robustness. Our implementation can be found at~\url{https://github.com/tmllab/2024_ICML_LAP}.

\end{abstract}
\section{Introduction}

Standard adversarial training (AT)~\cite{madry2018towards, zhang2019theoretically} is widely acknowledged as the most effective method for improving the robustness of deep neural networks (DNNs)~\cite{athalye2018obfuscated,croce2022evaluating}. Nevertheless, this training approach significantly increases the computational overhead due to the multi-step backward propagation, which limits its scalability for large networks and datasets. To alleviate this issue, several works~\cite{shafahi2019adversarial, wong2019fast, kim2021understanding} have introduced single-step AT as a time-efficient alternative, offering a balance between practicality and robustness.

\begin{figure}[t]
\begin{center}
\includegraphics[width=0.93\columnwidth]{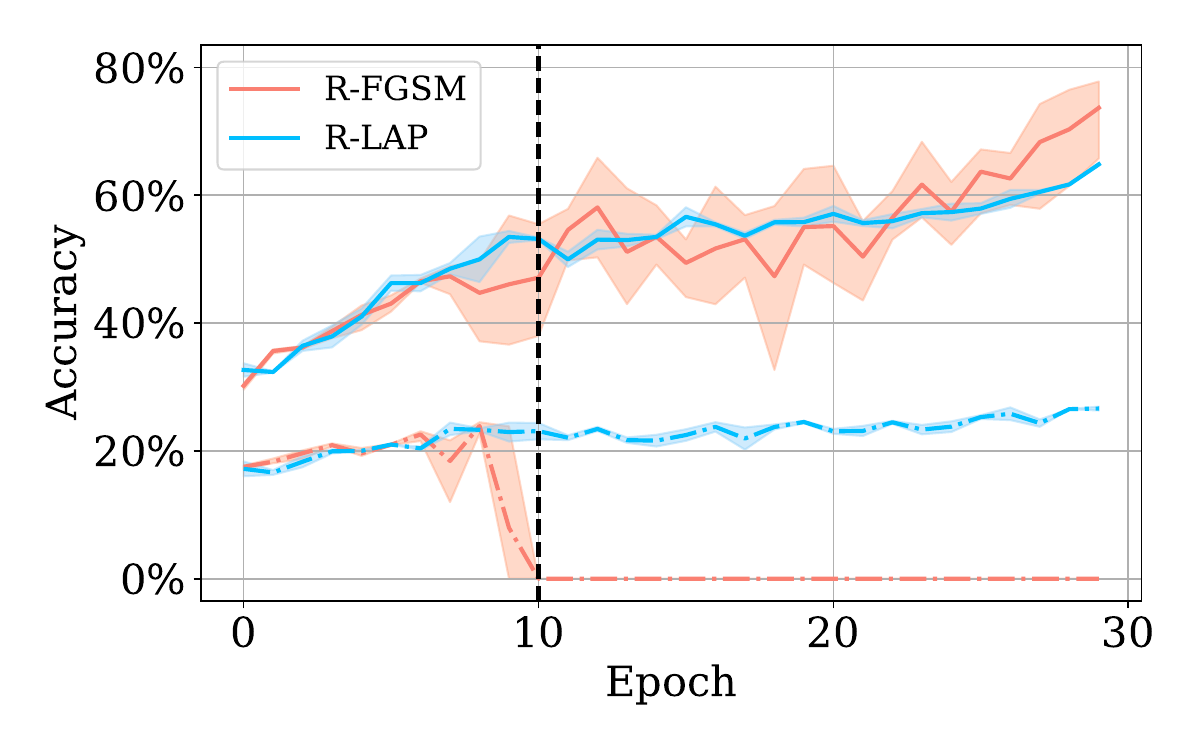}
\end{center}
\vspace{-1.6em}
\caption{The test accuracy of R-FGSM and R-LAP under 16/255 noise magnitude, where the solid and dashed lines denote natural and robust (PGD) accuracy, respectively.}
\vspace{-1.6em}
\label{fig:CO}
\end{figure}

Unfortunately, single-step AT faces a critical challenge known as catastrophic overfitting (CO)~\cite{wong2019fast}. This intriguing phenomenon is characterized by a sharp decline in the DNN's robustness, plummeting from peak to nearly zero in just a few iterations, as illustrated in Figure~\ref{fig:CO}. Prior studies~\cite{andriushchenko2020understanding, kim2021understanding} have pointed out that classifiers suffering from CO typically exhibit severely distorted decision boundaries. This distortion leads to a strange performance paradox in models affected by CO, as they can perfectly defend against single-step adversarial attacks but are highly vulnerable to multi-step adversarial attacks. However, the precise process of decision boundary distortion and the underlying factors that contribute to this performance paradox remain unclear.

To gain a detailed investigation of CO, we analyse the specific changes within individual DNN layers and their respective influences on the distortion of decision boundaries. More specifically, we identify the distinct transformations occurring in different layers during the CO process. The initial alterations in the DNN primarily occur in the former layers, leading to observable distorted decision boundaries and a subsequent reduction in robustness. As training progresses, each layer within DNNs experiences varying degrees of distortion. Notably, the former layers are more susceptible, showing markedly pronounced distortion, whereas the latter layers display relative resilience. As a result, forward propagation through these distorted former layers leads the model to exhibit sharp decision boundaries and manifest as CO.

Building on this, we delve into the underlying factors driving the transformation process that results in the distortion of decision boundaries and the performance paradox. Our research reveals that the heightened sensitivity in DNN's former layers can be attributed to the generation of \emph{pseudo-robust shortcuts}. These shortcuts, associated with certain large weights, empower the model to attain exceptional performance defence against single-step adversarial attacks. Nevertheless, relying solely on these shortcuts for decision-making induces the model to bypass genuine-robust learning, consequently distorting decision boundaries. By removing large weights from the former layers, we can effectively disrupt the improper reliance on these \emph{pseudo-robust shortcuts}, thereby gradually reinstating the robustness of DNNs in the CO state. The above analyses validate that the model's dependence on \emph{pseudo-robust shortcuts} for decision-making is the key factor triggering the occurrence of CO.

Motivated by these insights, our proposed method, \textbf{L}ayer-\textbf{A}ware Adversarial Weight \textbf{P}erturbation (LAP), is designed to prevent CO by hindering the generation of \emph{pseudo-robust shortcuts}. To realize this objective, LAP is strategically crafted to interrupt the model's stable reliance on these shortcuts by explicitly implementing adaptive weight perturbations across different layers. It is worth noting that our method simultaneously generates adversarial perturbations for both inputs and weights, thus avoiding any additional computational burden. We evaluate the effectiveness of our method across various adversarial attacks, datasets, and network architectures, showing that our proposed method can not only effectively eliminate CO but also further boost adversarial robustness, even under extreme noise magnitudes. Our main contributions are summarized as follows:

\begin{itemize}[leftmargin=14pt,topsep=1pt, itemsep=1pt]
\item We find that during CO, different layers undergo distinct changes, with the former layers exhibiting greater sensitivity, marked by earlier and more significant distortion.

\item We reveal that the generation and dependence on \emph{pseudo-robust shortcuts} trigger CO, which allows the model to precisely defend against single-step adversarial attacks but bypass genuine-robustness learning.

\item We propose the LAP method, which aims to obstruct the formation of \emph{pseudo-robust shortcuts}, thereby effectively preventing the occurrence of CO.
\end{itemize}
\section{Related Work}
In this section, we briefly review the relevant literature.
\subsection{Adversarial Training}
AT has been demonstrated to be the most effective defence method~\cite{athalye2018obfuscated, zhou2022modeling, dong2023competition} that is generally formulated as a min-max optimization problem~\cite{madry2018towards,croce2022evaluating, wang2024balance}, which is shown in the following formula:
\begin{equation}
\begin{aligned}
\min _{\mathbf{w}} \mathbb{E}_{\left\{\mathbf{x}_i, y_i\right\}_{i=1}^n}\left[\max _{\delta_i \in \epsilon_p} \ell(f_{\mathbf{w}}( x_i + \delta_i), y_i)\right],
\end{aligned}
\end{equation}
where $\{\mathbf{x}_i, y_i\}_{i=1}^n$ is the training dataset, $f$ is the classifier parameterized by $\mathbf{w}$, $\ell$ is the cross-entropy loss, $\delta$ is the perturbation confined within the $\epsilon$ radius $L_p$-norm ball.

Vanilla Fast Gradient Sign Method (V-FGSM)~\cite{goodfellow2014explaining} is a single-step maximization approach that utilizes one iteration to generate perturbations, defined as:
\begin{equation}
\begin{aligned}
\delta_{V-FGSM} = \epsilon \cdot \operatorname{sign}\left( \nabla_{x} \ell(f_{\mathbf{w}}(x_i),y_i)\right).
\end{aligned}
\end{equation}
Random FGSM (R-FGSM)~\cite{wong2019fast} and Noise FGSM (N-FGSM)~\cite{de2022make} adopt stronger noise initialization $(-\epsilon, \epsilon)$ and $(-2\epsilon, 2\epsilon)$, respectively, to further enhance the quality of maximization.

To improve robust generalization, Adversarial Weight Pertuabtion (AWP)~\cite{wu2020adversarial} introduces an extra weight perturbation process, which is formulated as follows:
\begin{equation}
\begin{aligned}
\min _{\mathbf{w}} \max _{\boldsymbol{\nu} \in \mathcal{V}} \frac{1}{n} \sum_{i=1}^{n} \max _{\delta_i \in \epsilon_p} \ell\left(f_{\mathbf{w}+\boldsymbol{\nu}}(x_{i} + \delta_i), y_{i})\right),
\end{aligned}
\label{eq:AWP}
\end{equation}
where $\mathcal{V}$ is a feasible region for the weight perturbation $\boldsymbol{\nu}$.

\subsection{Weight Perturbation}
The relationship between the geometry of the loss landscape and the model's generalization ability has been widely investigated~\cite{keskar2016large,dziugaite2017computing,huang2023robust,li2023understanding}. Recent works have demonstrated that random weight perturbations can effectively smooth the loss surface, thereby enhancing the generalization capacity~\cite{wen2018flipout,he2019parametric}. Building on this, several studies have utilized gradient information to generate adversarial weight perturbations, aiming to flatten the landscape in worst-case scenarios~\cite{ wu2020adversarial,foret2020sharpness,yu2022robust,yu2022understanding}. However, the impact of weight perturbation across different layers, as well as its role in CO, remains rarely explored.
\begin{figure*}[t]
\centering
\includegraphics[width=1.87\columnwidth]{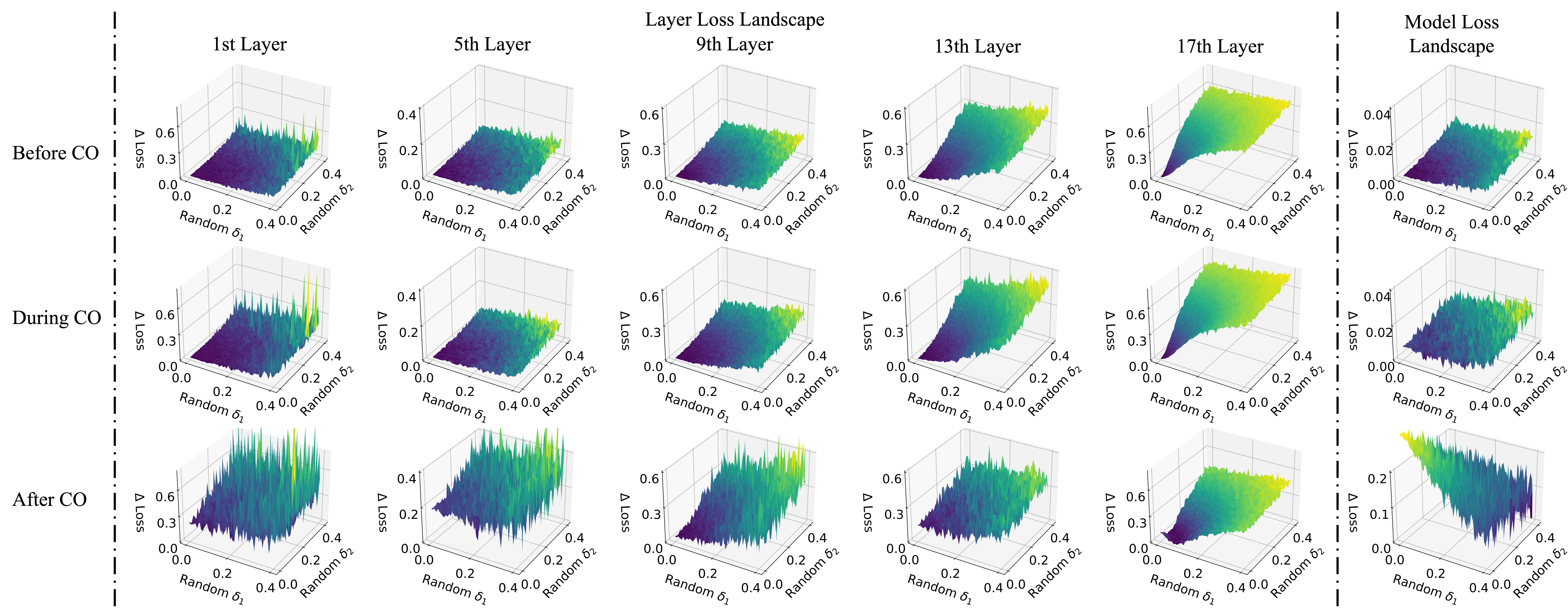}
\caption{Visualization of the loss landscape for individual layers (1st to 5th columns) and for the whole model (6th column). The upper, middle, and lower rows correspond to the stages before, during, and after CO, respectively.}
\vspace{-1.2em}
\label{fig:Landscape}
\end{figure*}
\subsection{Catastrophic Overfitting}
Since the identification of CO~\cite{wong2019fast}, a line of studies has been dedicated to understanding and addressing this intriguing phenomenon. \citet{de2022make, niu2022fast}~found that employing a stronger noise initialization can effectively delay the onset of CO. Additionally, \citet{andriushchenko2020understanding} observed that the models impacted by CO tend to become highly distorted and proposed a gradient align method to smooth local non-linear surfaces. Recent works have also introduced a variety of strategies designed to counter CO, including subspace extraction~\cite{li2022subspace}, gradient filtering~\cite{vivek2020single, golgooni2023zerograd,lin2023over}, adaptive perturbation~\cite{kim2021understanding, huang2023fast}, and local linearity~\cite{park2021reliably, sriramanan2021towards, lin2023eliminating, rocamora2023efficient}. Regrettably, the aforementioned methods either suffer from CO when faced with stronger adversaries or significantly increase the computational overhead. This study explores the changes within individual DNN layers and introduces a \emph{pseudo-robust shortcuts} dependency perspective, thereby proposing the LAP as an effective and efficient CO solution.

\begin{figure*}[t]
    \centering
        \begin{subfigure}
        {
            \includegraphics[width=0.397\columnwidth]{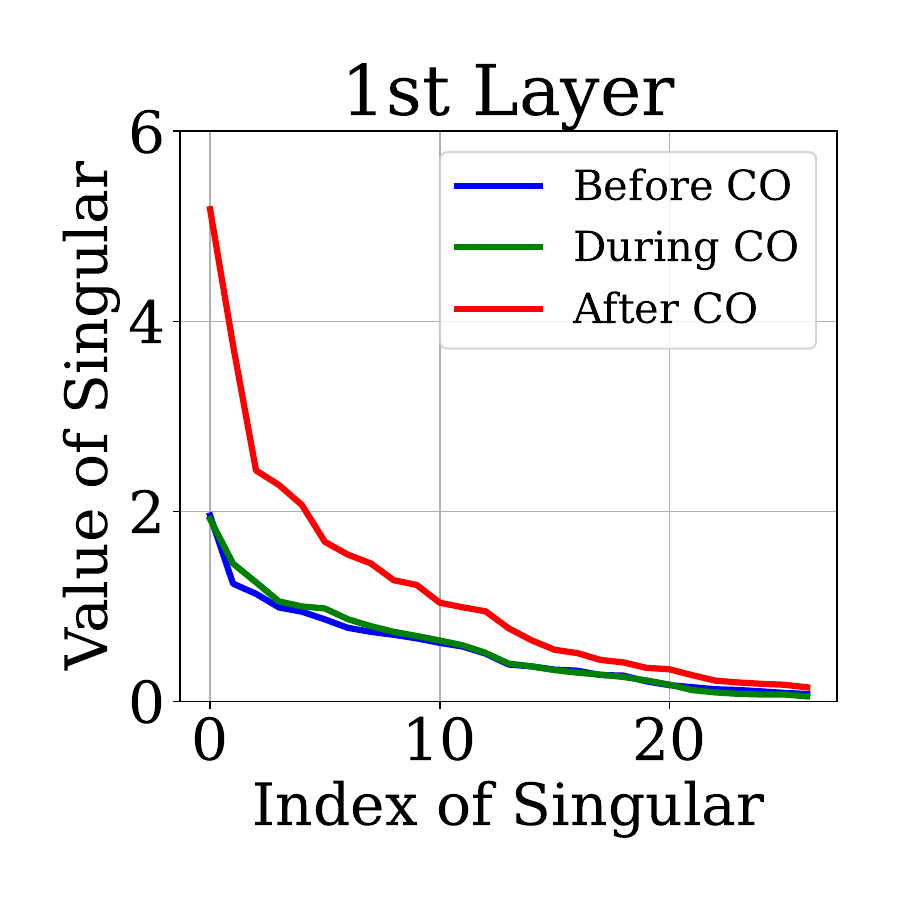}
            \includegraphics[width=0.397\columnwidth]{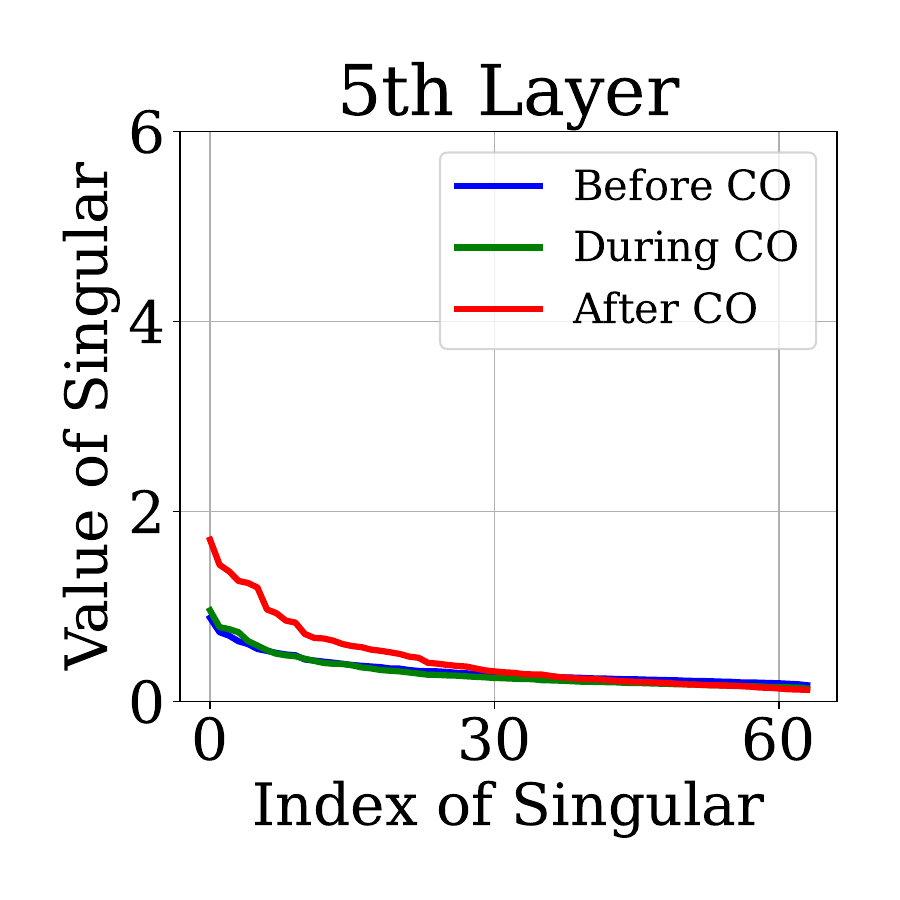}
            \includegraphics[width=0.397\columnwidth]{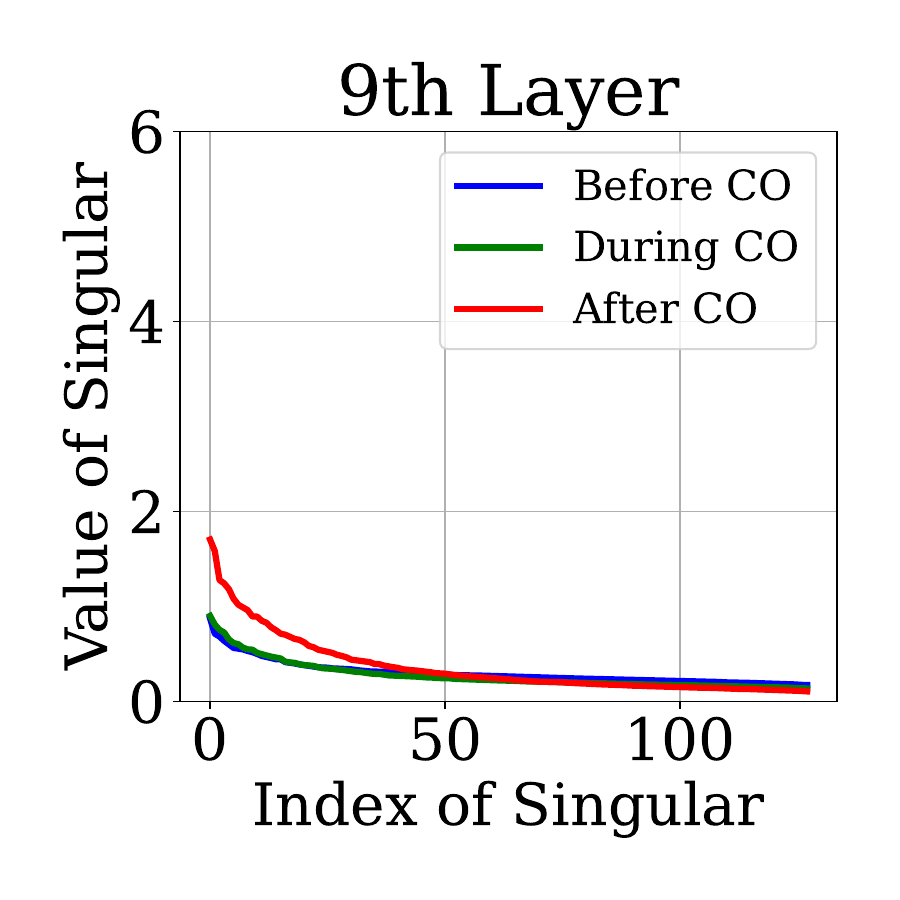}
            \includegraphics[width=0.397\columnwidth]{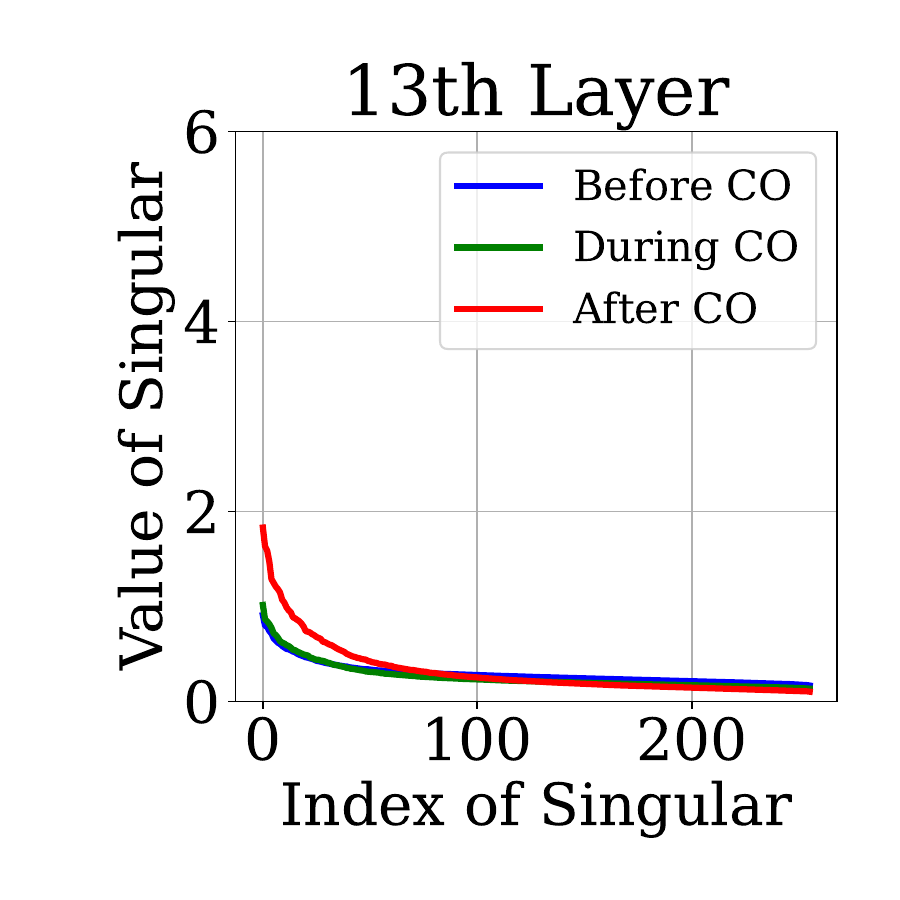}
            \includegraphics[width=0.397\columnwidth]{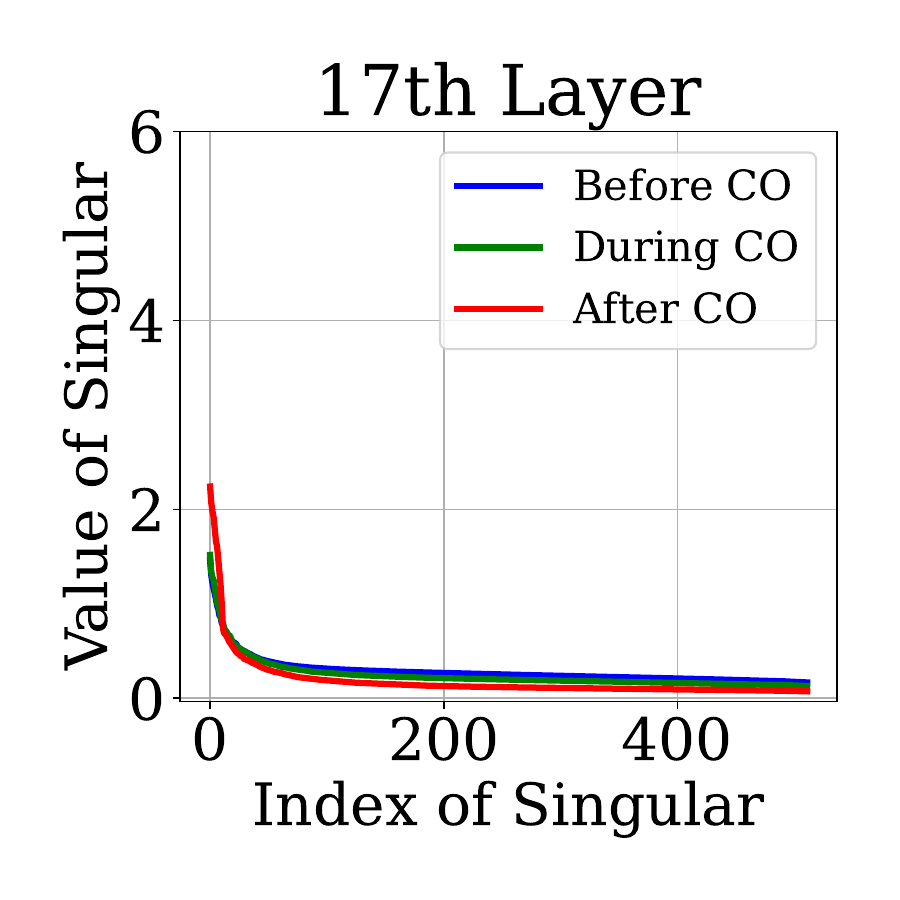}
        }
        \end{subfigure}
    \vspace{-1.2em}
    \caption{Singular value of weights (convolution kernel) at different DNN layers. The blue, green, and red lines represent the model state before, during, and after CO, respectively.}
    \vspace{-1.2em}
    \label{fig:Singular}
\end{figure*}

\section{Methodology}
In this section, we observe that during catastrophic overfitting (CO), different layers in deep neural networks (DNNs) undergo distinct changes, with the former layers being more prone to distortion (Section~\ref{section:3_1}). Subsequently, we reveal that the model's reliance on \emph{pseudo-robust shortcuts} for decision-making triggers CO (Section~\ref{section:3_2}). Consequently, we propose Layer-Aware Adversarial Weight Perturbation (LAP), which applies adaptive perturbations to eliminate the generation of shortcuts (Section~\ref{section:3_3}). Finally, we provide a theoretical analysis deriving an upper bound to ensure the effectiveness of our proposed method (Section~\ref{section:3_4}).

\subsection{Layers Transformation During CO}
\label{section:3_1}
Prior research~\cite{andriushchenko2020understanding, kim2021understanding} has shown that the decision boundaries of DNNs undergo significant distortion during the CO process, resulting in a performance paradox in response to single-step and multi-step adversarial attacks. Nevertheless, the prevailing studies on CO generally consider DNNs as a whole and focus on analysing the final output. However, considering an L-layer DNN with parameters $\{\mathbf{w}_{l}\}_{l=1}^L$, its output is an aggregation of forward propagation through these layers, denoted by $f_{\mathbf{w}}(x) = \mathbf{w}_L( \mathbf{w}_{L-1}\ldots(\mathbf{w}_{1}x)) $ for $l=1, \ldots, L$. Therefore, the specific impact of each layer on the distorted decision boundaries and the underlying factors that induce this performance paradox are still unclear.

In this work, we conduct a layer-by-layer investigation of the single-step AT throughout the training process, as illustrated in Figure~\ref{fig:Landscape}. Specifically, we utilize a PreActResNet-18 network trained on the CIFAR-10 dataset using the R-FGSM~\cite{wong2019fast} method under 16/255 noise magnitude. For visualizing the loss landscape of the whole model, we apply random perturbations to the input, denoted as $x + \delta$, and then compute the variation in loss, represented as $\Delta$ Loss. To analyse individual layers, we introduce random perturbations to the weights of the corresponding layer, expressed as $\mathbf{w}_l + \delta$ for $l=1, 5, 9, 13, 17$, and calculate the subsequent change in the loss.

As illustrated in Figure~\ref{fig:Landscape} (upper row), at the moment of peak robustness, both the whole model and its individual layers exhibit a flattened loss landscape. At this point, it becomes evident that the former layers display a higher degree of stability compared to the latter layers, as indicated by the smaller variations in loss due to the random perturbations.

With the onset of CO, the model manifests a decrease in robustness, accompanied by an observable distortion in the loss landscape, as illustrated in Figure~\ref{fig:Landscape} (middle row). The detailed analysis within each layer demonstrates that the former layers are the first to manifest increased sensitivity, characterized by a sharper loss landscape; in contrast, the latter layers undergo only minor transformations.

Following the occurrence of CO, the classifier's decision boundaries become completely distorted, rendering it extremely vulnerable to multi-step adversarial attacks, as depicted in Figure~\ref{fig:Landscape} (lower row). It can be observed that different layers exhibit distinct changes; the former layers experience the most severe distortion, marked by a significantly sharp surface, whereas the latter layers exhibit relative insensitivity. In summary, during the CO process, the former layers within DNNs undergo the most profound changes, transitioning from relatively stable to entirely sensitive.
 
\subsection{Pseudo-Robust Shortcut Dependency}
\label{section:3_2}
Subsequently, we delve into the underlying factors that induce the sensitivity transformation observed in DNNs during the CO process. To accomplish this objective, we examine the influence of weights on the model's decision-making process. In practice, we compute the singular values for each convolutional kernel to handle the extensive number of weights, as depicted in Figure~\ref{fig:Singular}. Before the CO occurrence, a fairly uniform distribution of singular values is observed across all layers. However, after CO, there is a noticeable increase in the variance of singular values, leading to sharper model output, as discussed in Section~\ref{section:3_1}. This significant rise in large singular values suggests the growing importance of certain weights in the model's decision-making. Remarkably, the former layers exhibit the most pronounced increase in large singular values, nearly tripling from before, indicating that the model's decision-making becomes heavily dependent on certain weights in these layers.
 
\begin{figure*}[t]
    \centering
        \begin{subfigure}[Remove random weights, small weights, and large weights from the former (1st to 5th) layers, as shown in the left, middle, and right columns, respectively.]{
            \includegraphics[width=0.495\columnwidth]{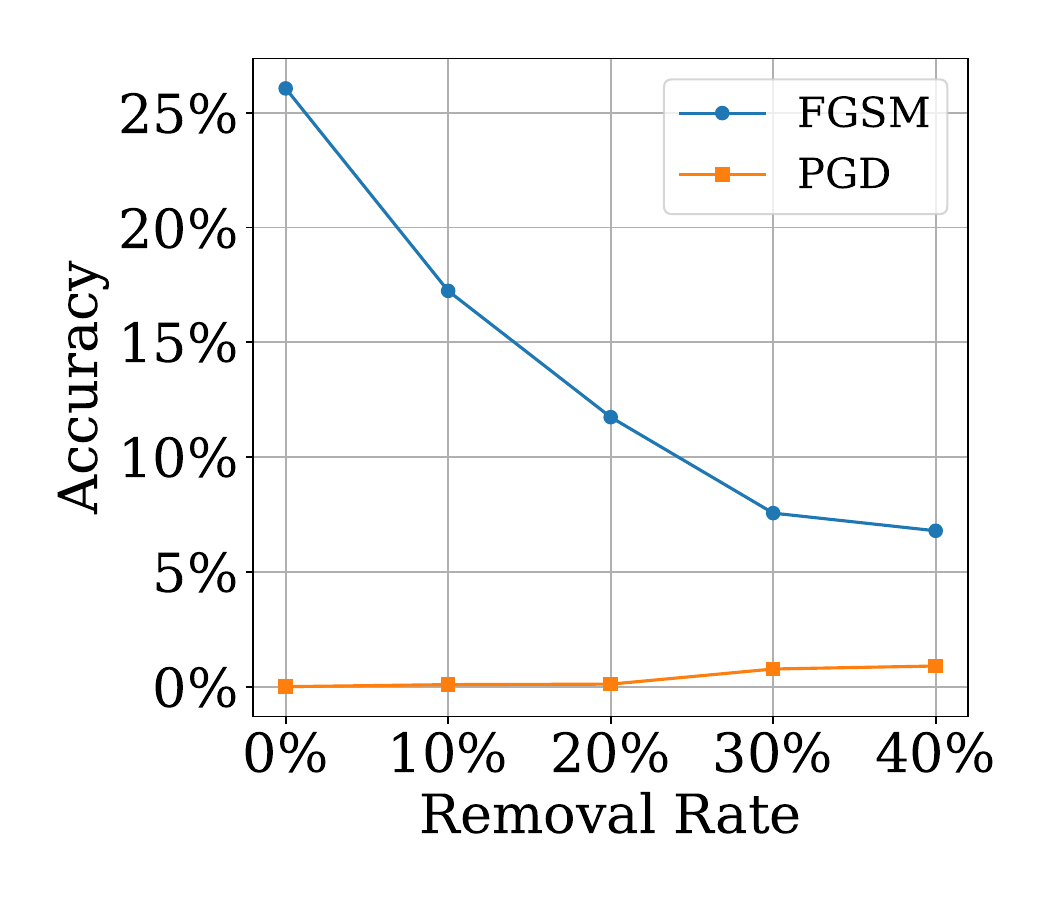}
            \includegraphics[width=0.495\columnwidth]{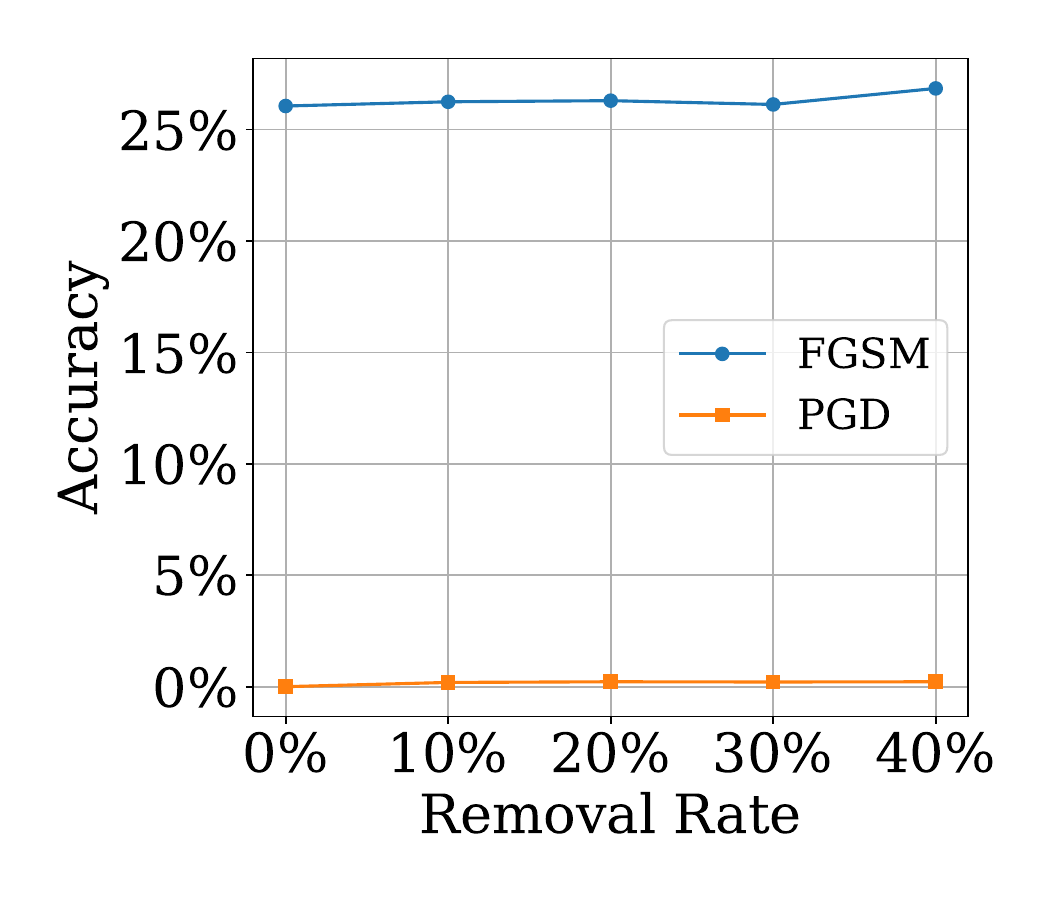}
            \includegraphics[width=0.495\columnwidth]{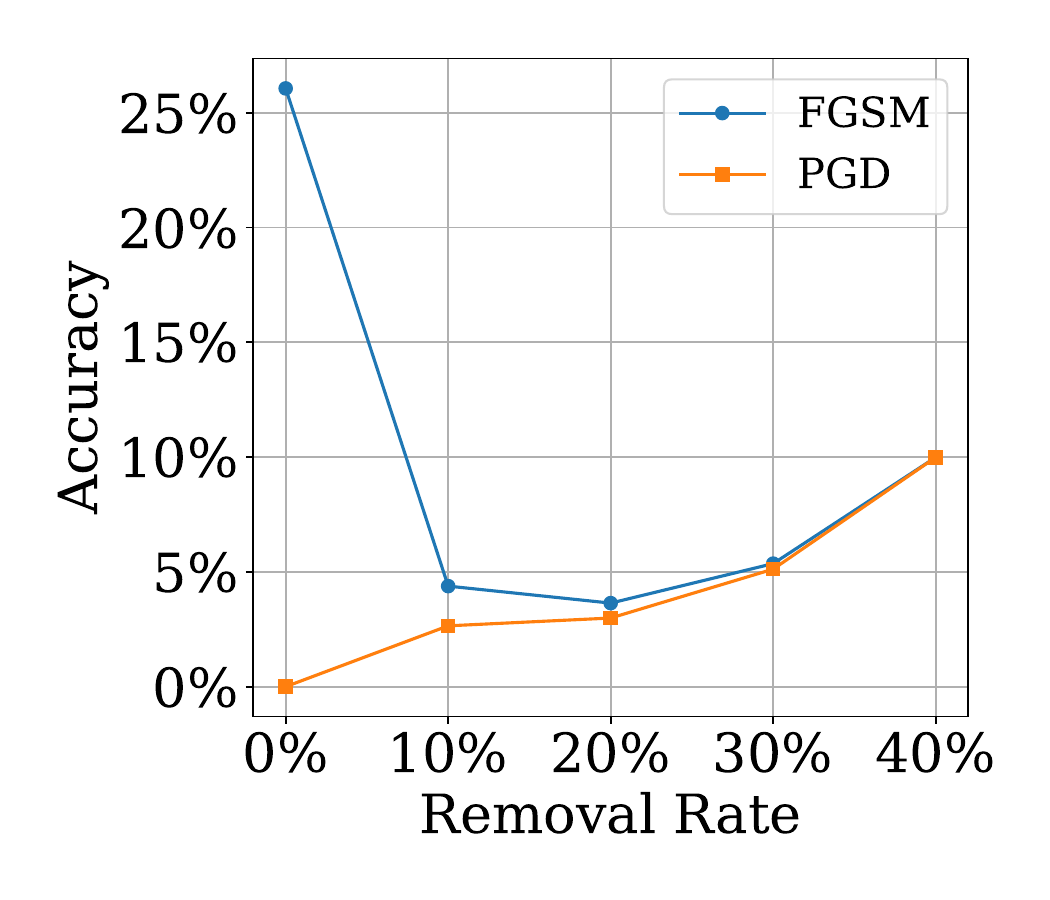}
            \label{fig:RemovalA}}  
        \end{subfigure}
        \centering
        \begin{subfigure}[Remove large weights from the latter (14th to 17th) layers.]{
            \includegraphics[width=0.495\columnwidth]{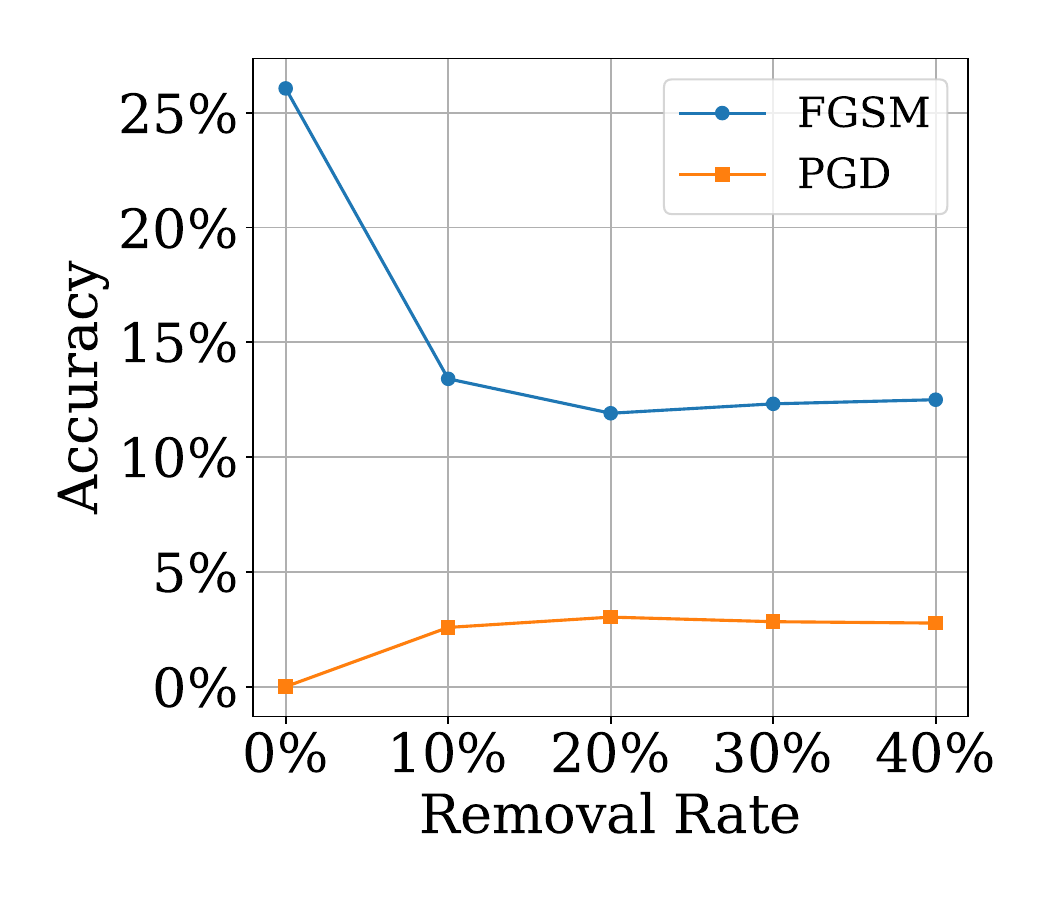}
            \label{fig:RemovalB}}
        \end{subfigure}
    \vspace{-0.8em}
    \caption{Evaluating the test accuracy of a CO-affected model against single-step (FGSM) and multi-step (PGD) adversarial attack.}
    \label{fig:Removal}
    \vspace{-0.8em}
\end{figure*}

In order to gain deeper insight into this dependency, we randomly removed some weights from the former (1st to 5th) layers in a model already affected by CO, as illustrated in Figure~\ref{fig:RemovalA} (left column). With the increased removal rate, the model's accuracy under the FGSM attack decreased from 26\% to 6\%, whereas its accuracy against the PGD attack showed a slight increase. This anomalous trend indicates a performance paradox in models impacted by CO under FGSM and PGD attacks, contrasting with genuine-robust models where higher FGSM accuracy generally implies greater PGD accuracy. Therefore, we propose that the heightened sensitivity in the former layers originates from the generation of \emph{pseudo-robust shortcuts}, solely relying on them can effectively defend against single-step adversarial attacks but bypass genuine-robust learning.

To further substantiate our perspective, we investigate the particular weights associated with these \emph{pseudo-robust shortcuts}. As shown in Figure~\ref{fig:RemovalA} (middle column), the removal of small weights in the former layers has a negligible impact on the model's performance against both FGSM and PGD attacks, suggesting a weak relevance between these weights and shortcuts. Conversely, removing only 10\% of the large weights can effectively interrupt the \emph{pseudo-robust shortcuts}, resulting in a notable 22\% reduction in FGSM attack accuracy and reinstatement of robustness against PGD attack to 2.65\%, as depicted in Figure~\ref{fig:RemovalA} (right column). With the gradual removal of larger weights, the model not only shows an improvement in robustness but also successfully overcomes the performance paradox against FGSM and PGD attacks. For a fair comparison, we also remove the large weights from the latter (14th to 17th) layers, as depicted in Figure~\ref{fig:RemovalB}. Clearly, the same intervention in the latter layers is less effective, highlighting the \emph{pseudo-robust shortcuts} that play a critical role in the CO phenomenon, primarily present in the former layer.

Conclusively, we introduce the perspective of \emph{pseudo-robust shortcuts} dependency to explain the occurrence of CO. Specifically, the heightened sensitivity of DNN can be attributed to its decision-making solely dependent on \emph{pseudo-robust shortcuts}, which are typically associated with certain large weights in former layers. These shortcuts, although exceptionally accurate in defending against single-step adversarial attacks, induce the model to bypass genuine-robust learning, thereby resulting in distorted decision boundaries and triggering the performance paradox in CO.

\subsection{Proposed Method}
\label{section:3_3}
Building upon our perspective, our objective is to eliminate the formation of \emph{pseudo-robust shortcuts}, thereby effectively preventing the occurrence of CO. Inspired by AWP \cite{wu2020adversarial} and SAM~\cite{foret2020sharpness}, we introduce the \textbf{L}ayer-\textbf{A}ware Adversarial Weight \textbf{P}erturbation (LAP) method that explicitly implements adaptive weight perturbations across different layers to hinder the generation of \emph{pseudo-robust shortcuts}, which can be expressed as follows:
\begin{equation}
\begin{aligned}
\min _{\mathbf{w}} \frac{1}{n} \sum_{i=1}^{n}  \max _{\delta_i} \max _{\boldsymbol{\nu}_{l}}\ell\left(f_{\mathbf{w}+\boldsymbol{\nu}_{l}}(x_{i} + \delta_i), y_{i})\right).
\end{aligned}
\label{eq:LAP-A}
\end{equation}

To closely align with our goal, we introduce three novel improvements. Firstly, our method accumulates weight perturbations to effectively break persistent shortcuts by maintaining a larger magnitude of alteration. Secondly, we prioritize generating weight perturbations over input perturbations, aiming to obstruct the model from establishing stable shortcuts between inputs and weights. Thirdly, recognizing the distinct transformations in each layer, our approach adopts a gradually decreasing weight perturbation strategy from the former to the latter layer to avoid unnecessary redundant perturbations, as summarized below:
\begin{equation}
\begin{aligned}
\lambda_l = \beta \cdot \left( 1 - \left(\frac{\ln(l)}{\ln(L + 1)}\right)^\gamma \right), \,\,\, \text{for } l = 1, \ldots, L
\end{aligned}
\end{equation}
where $\lambda_l$ is the layer-aware perturbation, $\beta$ is the step size, and $\gamma$ controls the different layers strength.

However, the above design still requires extra backward propagation, which diminishes the efficiency advantage of single-step AT. To address this issue, we propose an efficient LAP implementation that concurrently generates adversarial perturbations for both inputs and weights, as detailed below:
\begin{equation}
\begin{aligned}
\min _{\mathbf{w}} \frac{1}{n} \sum_{i=1}^{n}  \max _{\delta_i, \boldsymbol{\nu}_{l}} \ell\left(f_{\mathbf{w}+\boldsymbol{\nu}_{l}}(x_{i} + \delta_i), y_{i})\right).
\end{aligned}
\label{eq:LAP}
\end{equation}
We further elucidate the intuitive basis for the efficient implementation of LAP. For a given input, its corresponding adversarial perturbation is generated by maximizing the loss value, which is calculated from both the network weights and the loss function. Assuming the loss function is Lipschitz continuous with a constant $\mathbb{L}$, the change in loss due to weight perturbation can be bounded as follows:
\begin{equation}
\fontsize{8.5}{12.0}\selectfont
\begin{aligned}
\left|\ell\left(f_{\mathbf{w}+\nu_l}(x), y\right)-\ell\left(f_{\mathrm{w}}(x), y\right)\right| \leq \mathbb{L}\left\|f_{\mathbf{w}+\nu_l}(x)-f_{\mathbf{w}}(x)\right\|.
\end{aligned}
\end{equation}
Hence, the variation in loss value is directly related to the changes in the model's output, which results from the aggregation of multiple layers, as outlined below:
\begin{equation}
\fontsize{9.0}{12.0}\selectfont
\begin{aligned}
f_{\mathbf{w}+\boldsymbol{\nu}_{l}}(x) - f_{\mathbf{w}}(x)  = \prod_{l=1}^L (\mathbf{w}_l +\boldsymbol{\nu}_l) \cdot x - \prod_{l=1}^L (\mathbf{w}_l) \cdot x.
\end{aligned}
\end{equation}
The above analysis reveals a positive correlation between changes in output and the magnitudes of weight perturbations. In practice, we employ a small weight perturbation size to restrict this magnitude. Meanwhile, our optimization objective is to attain a flattened weight loss landscape, ensuring that the introduction of small weight perturbations leads to relatively minor alterations in the loss value. Therefore, this discussion empirically demonstrates that the input perturbation, generated based on the original weights, has a high probability of retaining its effectiveness after the injection of weight perturbations, consequently enabling us to simultaneously generate both input and weight perturbations. The LAP algorithm is summarized in Algorithm~\ref{alg:algorithm}.

\begin{algorithm}[t]
	\renewcommand{\algorithmicrequire}{\textbf{Input:}}
	\renewcommand{\algorithmicensure}{\textbf{Output:}}
	\caption{ \emph{Layer-Aware Adversarial Weight Perturbation}}
	\label{alg:algorithm}
	\begin{algorithmic}[1]
		\REQUIRE L-layer Network $f_{\mathbf{w}}$, training data $\{\mathbf{x}_i, y_i\}_{i=1}^n$, training epoch $T$, batch size $N$, input perturbation size $\alpha$, layer-aware weight perturbation size $\lambda_l$.
            \ENSURE Adversarially robust model $f_{\mathbf{w}}$.
		\FOR{$t=1 \ldots T$ to} 
		\FOR{$i=1 \ldots N$ to}
            \STATE \# simultaneously generate $\delta_i$ and $\boldsymbol{\nu}_{l}$.
		\STATE $\delta_i = \alpha \cdot \operatorname{sign}\left(\nabla_{x} \ell(f_{\mathbf{w}}(x_{i}), y_{i})\right)$ 
            \STATE $\boldsymbol{\nu}_{l} = \lambda_l \cdot \frac{\nabla_{\mathbf{w}} \ell\left(f_{\mathbf{w}}(x_{i}), y_{i})\right)} {\left\|\nabla_{\mathbf{w}} \ell\left(f_{\mathbf{w}}(x_{i}), y_{i})\right)\right\|}\|\mathbf{w}\|$ 
		\STATE $LAP = \frac{1}{n} \sum_{i=1}^{n} \ell\left(f_{\mathbf{w}+\boldsymbol{\nu}_{l}}(x_{i} + \delta_i), y_{i})\right)$   
		\STATE $\mathbf{w} =(\mathbf{w}+\boldsymbol{\nu}_{l} )-\nabla_{\mathbf{w}+\boldsymbol{\nu}_{l}} \left(LAP\right)$
		\ENDFOR      
		\ENDFOR  
	\end{algorithmic}
\end{algorithm}

\begin{table*}[t]
\fontsize{7.7}{12.0}\selectfont
\caption{Comparison of CIFAR-10 test accuracy (\%) for various methods under different noise magnitudes. The results are averaged over three random seeds and reported with the standard deviation.}
\vspace{-1.2em}
\label{tab:CIFAR10}
  \begin{center}
  \begin{tabular}{l c c c c c c c c c c c c c }
    \toprule
    \toprule
    \multirow{2}*{Method} & \multicolumn{2}{c}{8/255} && \multicolumn{2}{c}{12/255} && \multicolumn{2}{c}{16/255} && \multicolumn{2}{c}{32/255} \\
    \cmidrule{2-3}
    \cmidrule{5-6}
    \cmidrule{8-9}
    \cmidrule{11-12}
    & Natural  & Auto Attack  && Natural  & Auto Attack  && Natural  & Auto Attack  && Natural  & Auto Attack\\
    \midrule
    \multirow{1}*{AT Free}& 76.52{\scriptsize$\pm$1.34} & 40.13{\scriptsize$\pm$0.39} && 68.28{\scriptsize$\pm$0.13} & 27.65{\scriptsize$\pm$0.38} && \,\,\,55.91{\scriptsize$\pm$10.94} & \,\,\,0.00{\scriptsize$\pm$0.00} && \,\,\,59.25{\scriptsize$\pm$10.98} & \,\,\,0.00{\scriptsize$\pm$0.00}\\
    \multirow{1}*{Grad Align}& 82.35{\scriptsize$\pm$0.92} & 44.76{\scriptsize$\pm$0.02} && 74.80{\scriptsize$\pm$0.64} & 29.88{\scriptsize$\pm$0.23} && 61.10{\scriptsize$\pm$0.49} & 19.07{\scriptsize$\pm$0.28} && 24.15{\scriptsize$\pm$4.03} & \,\,\,6.71{\scriptsize$\pm$2.31}\\
    \multirow{1}*{ZeroGrad} &81.71{\scriptsize$\pm$0.21} & 43.28{\scriptsize$\pm$0.18} && 77.75{\scriptsize$\pm$0.20}  & 22.56{\scriptsize$\pm$0.05} && 82.54{\scriptsize$\pm$0.19} & \,\,\,0.00{\scriptsize$\pm$0.00} && 68.95{\scriptsize$\pm$2.51} & \,\,\,0.00{\scriptsize$\pm$0.00}\\
    \multirow{1}*{MultiGrad} &81.83{\scriptsize$\pm$0.31} & 44.19{\scriptsize$\pm$0.10} && 83.72{\scriptsize$\pm$1.47} & \,\,\,0.00{\scriptsize$\pm$0.00} && 81.59{\scriptsize$\pm$3.19} & \,\,\,0.00{\scriptsize$\pm$0.00} && 73.50{\scriptsize$\pm$4.90} & \,\,\,0.00{\scriptsize$\pm$0.00}\\
    \multirow{1}*{V-FGSM}&84.26{\scriptsize$\pm$4.18} & \,\,\,0.00{\scriptsize$\pm$0.00} && 79.92{\scriptsize$\pm$1.82} & \,\,\,0.00{\scriptsize$\pm$0.00} && 72.72{\scriptsize$\pm$3.04} & \,\,\,0.00{\scriptsize$\pm$0.00} && 65.52{\scriptsize$\pm$2.15} & \,\,\,0.00{\scriptsize$\pm$0.00}\\
    \rowcolor{LightGreen}
    \multirow{1}*{V-LAP} &79.09{\scriptsize$\pm$0.78} & 41.24{\scriptsize$\pm$0.51} && 66.20{\scriptsize$\pm$0.42} & 24.07{\scriptsize$\pm$0.34} && 56.02{\scriptsize$\pm$0.07} & 15.17{\scriptsize$\pm$0.31} && 17.76{\scriptsize$\pm$3.11} & \,\,\,7.12{\scriptsize$\pm$0.64}\\
    \multirow{1}*{R-FGSM} &84.12{\scriptsize$\pm$0.29} & 42.88{\scriptsize$\pm$0.09} && 79.49{\scriptsize$\pm$4.57} & \,\,\,0.00{\scriptsize$\pm$0.00} && 73.67{\scriptsize$\pm$6.86} & \,\,\,0.00{\scriptsize$\pm$0.00} && 33.31{\scriptsize$\pm$8.31} & \,\,\,0.00{\scriptsize$\pm$0.00}\\
    \rowcolor{LightGreen}
    \multirow{1}*{R-LAP}& 83.81{\scriptsize$\pm$0.24} & 43.14{\scriptsize$\pm$0.45} && 74.10{\scriptsize$\pm$0.31} & 26.04{\scriptsize$\pm$1.04} && 64.83{\scriptsize$\pm$0.29} & 15.69{\scriptsize$\pm$0.28} && 27.49{\scriptsize$\pm$0.48} & \,\,\,8.04{\scriptsize$\pm$0.63}\\
    \multirow{1}*{N-FGSM}& 80.40{\scriptsize$\pm$0.16} & 44.21{\scriptsize$\pm$0.47} && 71.44{\scriptsize$\pm$0.16} & 30.25{\scriptsize$\pm$0.06} && 62.91{\scriptsize$\pm$1.03} & 18.58{\scriptsize$\pm$2.28} && 27.66{\scriptsize$\pm$3.57} & \,\,\,0.00{\scriptsize$\pm$0.00}\\
    \rowcolor{LightGreen}
    \multirow{1}*{N-LAP} &80.76{\scriptsize$\pm$0.15} & \textbf{44.97}{\scriptsize$\pm$0.24} && 71.91{\scriptsize$\pm$0.19} & \textbf{30.60}{\scriptsize$\pm$0.27} && 63.73{\scriptsize$\pm$0.27} & \textbf{19.55}{\scriptsize$\pm$0.18} && 29.19{\scriptsize$\pm$1.00} & \,\,\,\textbf{8.85}{\scriptsize$\pm$1.48}\\
    \midrule
    \multirow{1}*{PGD-2} &84.72{\scriptsize$\pm$0.08} & 42.92{\scriptsize$\pm$0.60} && 79.13{\scriptsize$\pm$0.25} & 28.30{\scriptsize$\pm$0.35} && 72.50{\scriptsize$\pm$0.51} & 17.89{\scriptsize$\pm$0.16} && 48.99{\scriptsize$\pm$0.19} & \,\,\,3.76{\scriptsize$\pm$0.02}\\
    \multirow{1}*{PGD-10} &80.91{\scriptsize$\pm$0.52} & \textbf{46.37}{\scriptsize$\pm$0.76} && 72.03{\scriptsize$\pm$0.30} & \textbf{33.13}{\scriptsize$\pm$0.28} && 67.61{\scriptsize$\pm$0.83} & \textbf{21.98}{\scriptsize$\pm$0.30} && 35.28{\scriptsize$\pm$0.78} & \textbf{10.88}{\scriptsize$\pm$0.41}\\
    \bottomrule
    \bottomrule
  \end{tabular}
  \end{center}
\vspace{-1.2em}
\end{table*}

\subsection{Theoretical Analysis}
\label{section:3_4}
Furthermore, we provide a theoretical analysis to derive an upper bound on the expected error of our method. Building upon the previous PAC-Bayesian framework~\cite{neyshabur2017exploring, wu2020adversarial} and assuming a prior distribution $\mathbb{P} \sim \mathcal N (0, \sigma^2)$ for the weights, we can formulate the upper bound for the expected error of the classifier, with a probability of at least $1 - \delta$ across the $n$ training samples:

\begin{equation}
\begin{aligned}
\mathbb{E}_{\boldsymbol{\nu}}\left[\ell \left(f_{\mathbf{w}+\boldsymbol{\nu}}\right)\right] & \leq {\mathbb{E}_{\boldsymbol{\nu}}\left[\hat{\ell}\left(f_{\mathbf{w}+\boldsymbol{\nu}}\right)\right]} \\
&+ 4 \sqrt{\frac{1}{n}\left(K L(\mathbf{w}+\boldsymbol{\nu} \| P)+\ln \frac{2n}{\delta}\right)}.
\end{aligned}
\end{equation}

Considering the weight perturbation in the worst-case scenario $ max_{\boldsymbol{\nu}}[\hat{\ell} \left(f_{\mathbf{w}+\boldsymbol{\nu}}\right)]$, and the standard deviation of the weight perturbation relation to the layer magnitude $\sigma_l=\lambda_l \cdot \|\mathbf{w}_l\|_2$, the PAC-Bayes bound of our proposed LAP method can be controlled as follows:
\begin{equation}
\begin{aligned}
    \mathbb{E}_{\left\{\mathbf{x}_i, y_i\right\}_{i=1}^n, \{\boldsymbol{\nu}_l\}_{l=1}^L}  &[\ell\left(f_{\mathbf{w} + \boldsymbol{\nu}_l}\right)] \leq \hat{\ell}\left(f_{\mathbf{w}}\right) \\
    & \! \! \! \! \! \! \! \! \! \! \! \! \! \! \! \! \! + \left\{max_{\{\boldsymbol{\nu}_l\}_{l=1}^L}[\hat{\ell}\left(f_{\mathbf{w} + \boldsymbol{\nu}_l}\right)] - \hat{\ell}\left(f_{\mathbf{w}}\right)\right\} \\
    & \! \! \! \! \! \! \! \! \! \! \! \! \! \! \! \! \! + 4 \sqrt{\frac{1}{n} \left(\sum_{l=1}^L \left(\frac{1}{2\lambda_l^2}\right) + \ln \frac{2 n}{\delta}\right)}.
\end{aligned}
\end{equation}
\section{Experiment}

In this section, we evaluate the effectiveness of LAP, including experiment settings (Section~\ref{section:4_1}), performance evaluations (Section~\ref{section:4_2}), ablation studies (Section~\ref{section:4_3}), and training cost analysis (Section~\ref{section:4_4}).

\subsection{Experiment Setting}
\label{section:4_1}

\begin{table*}[t]

\fontsize{8.2}{12.0}\selectfont
\caption{Comparison of CIFAR-100 test accuracy (\%) for various methods under different noise magnitudes. The results are averaged over three random seeds and reported with the standard deviation.}
\label{tab:CIFAR100}
  \begin{center}
  \begin{tabular}{l c c c c c c c c c c c c c }
    \toprule
    \toprule
    \multirow{2}*{Method} & \multicolumn{2}{c}{8/255} && \multicolumn{2}{c}{12/255} && \multicolumn{2}{c}{16/255} && \multicolumn{2}{c}{32/255} \\
    \cmidrule{2-3}
    \cmidrule{5-6}
    \cmidrule{8-9}
    \cmidrule{11-12}
    & Natural  & Auto Attack  && Natural  & Auto Attack  && Natural  & Auto Attack  && Natural  & Auto Attack \\
    \midrule
    \multirow{1}*{V-FGSM}& 54.87{\scriptsize$\pm$2.53} & \,\,\,0.00{\scriptsize$\pm$0.00} && 45.40{\scriptsize$\pm$1.89} & \,\,\,0.00{\scriptsize$\pm$0.00} && 41.38{\scriptsize$\pm$6.03} & \,\,\,0.00{\scriptsize$\pm$0.00} && 27.22{\scriptsize$\pm$4.54} & 0.00{\scriptsize$\pm$0.00}\\
    \rowcolor{LightGreen}
    \multirow{1}*{V-LAP}& 53.07{\scriptsize$\pm$0.59} & 19.49{\scriptsize$\pm$0.57} && 42.24{\scriptsize$\pm$0.29} & 11.27{\scriptsize$\pm$0.33} && 34.30{\scriptsize$\pm$0.19} & \,\,\,7.63{\scriptsize$\pm$0.52} && \,\,\,9.37{\scriptsize$\pm$1.76} & 1.33{\scriptsize$\pm$0.21}\\
    \multirow{1}*{R-FGSM}& 60.29{\scriptsize$\pm$2.12} & \,\,\,0.00{\scriptsize$\pm$0.00} && 21.18{\scriptsize$\pm$9.56} & \,\,\,0.00{\scriptsize$\pm$0.00} && 11.46{\scriptsize$\pm$7.33} & \,\,\,0.00{\scriptsize$\pm$0.00} && 13.56{\scriptsize$\pm$10.95} & 0.00{\scriptsize$\pm$0.00}\\
    \rowcolor{LightGreen}
    \multirow{1}*{R-LAP}&58.75{\scriptsize$\pm$0.20} & 21.62{\scriptsize$\pm$0.01} && 49.74{\scriptsize$\pm$0.29} & 12.10{\scriptsize$\pm$0.13} && 39.13{\scriptsize$\pm$0.46} & \,\,\,7.98{\scriptsize$\pm$1.09} && 19.52{\scriptsize$\pm$0.84} & 2.50{\scriptsize$\pm$0.44}\\
    \multirow{1}*{N-FGSM}&55.19{\scriptsize$\pm$0.35} & 22.46{\scriptsize$\pm$0.12} && 46.16{\scriptsize$\pm$0.18} & 14.51{\scriptsize$\pm$0.11} && 37.71{\scriptsize$\pm$0.06} & 10.22{\scriptsize$\pm$0.18} && 18.29{\scriptsize$\pm$5.64} & 0.00{\scriptsize$\pm$0.00}\\
    \rowcolor{LightGreen}
    \multirow{1}*{N-LAP}&55.12{\scriptsize$\pm$0.20} & \textbf{23.15}{\scriptsize$\pm$0.28} && 46.76{\scriptsize$\pm$0.18} & \textbf{15.16}{\scriptsize$\pm$0.04} && 38.02{\scriptsize$\pm$0.11} & \textbf{10.40}{\scriptsize$\pm$0.14} && 16.85{\scriptsize$\pm$0.83} & \textbf{3.45}{\scriptsize$\pm$0.28}\\
    \midrule
    \multirow{1}*{PGD-2}&60.09{\scriptsize$\pm$0.20} & 22.52{\scriptsize$\pm$0.14} && 53.46{\scriptsize$\pm$0.27} & 13.69{\scriptsize$\pm$0.02} && 47.50{\scriptsize$\pm$0.28} & \,\,\,9.56{\scriptsize$\pm$0.07} && 31.89{\scriptsize$\pm$0.69} & 1.76{\scriptsize$\pm$0.22}\\
    \multirow{1}*{PGD-10}&55.20{\scriptsize$\pm$0.31} & \textbf{23.71}{\scriptsize$\pm$0.11} && 47.74{\scriptsize$\pm$0.15} & \textbf{15.52}{\scriptsize$\pm$0.06} && 42.21{\scriptsize$\pm$0.16} & \textbf{10.87}{\scriptsize$\pm$0.07} && 21.82{\scriptsize$\pm$0.21} & \textbf{4.03}{\scriptsize$\pm$0.08}\\
    \bottomrule
    \bottomrule
  \end{tabular}
  \end{center}
\vspace{-1.0em}
\end{table*}

\begin{figure*}[t]
\centering
\includegraphics[width=2.00\columnwidth]{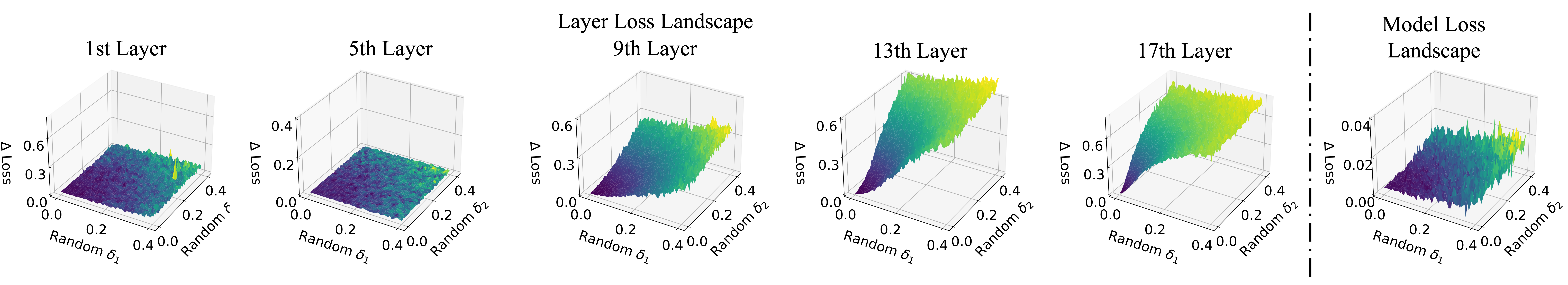}
\vspace{-0.4em}
\caption{Visualization of the loss landscape for individual layers (1st to 5th columns) and for the whole model (6th column).}
\vspace{-1.5em}
\label{fig:Landscapeour}
\end{figure*}

\textbf{Baselines.}\hspace*{2mm}We select a range of popular single-step AT methods for compare with LAP, which includes V-FGSM \cite{goodfellow2014explaining}, R-FGSM~\cite{wong2019fast}, N-FGSM~\cite{de2022make}, FreeAT~\cite{shafahi2019adversarial}, Grad Align~\cite{andriushchenko2020understanding}, ZeroGrad and MultiGrad~\cite{golgooni2023zerograd}. Additionally, we present the results of the iterative-step AT method PGD-2 and PGD-10~\cite{madry2018towards} as a reference for ideal performance.

\vspace{0.2em}
\textbf{Datasets and Model Architectures.}\hspace*{2mm}We use three benchmark datasets, CIFAR-10, CIFAR-100~\cite{krizhevsky2009learning} and Tiny-ImageNet~\cite{netzer2011reading}, for evaluating the performances of our proposed method. The widely-used data augmentation random cropping and horizontal flipping are applied to these datasets. The settings and results on Tiny-ImageNet can be found in Appendix~\ref{appendix:B}. For a comprehensive evaluation, we report the training from scratch results on PreActResNet-18~\cite{he2016identity}, WideResNet-34~\cite{zagoruyko2016wide}, and Vit-small~\cite{dosovitskiy2020image} architectures. The results of WideResNet-34 and Vit-small are provided in Appendix~\ref{appendix:A}.

\vspace{0.2em}
\textbf{Learning Rate Schedule.}\hspace*{2mm}We use the cyclical learning rate schedule~\cite{smith2017cyclical} spanning 30 epochs, which reaches its maximum learning rate of 0.2 at the 15th epoch. The results of the piecewise learning rate schedule with 200 training epochs are available in Appendix~\ref{appendix:C}.

\vspace{0.2em}
\textbf{Adversarial Evaluation.}\hspace*{2mm}In order to thoroughly assess the models' robustness, we utilize the widely-used PGD attack configuration with 50 steps and 10 restarts~\cite{wong2019fast}, as well as the Auto Attack~\cite{croce2020reliable}.

\textbf{Setup for LAP.}\hspace*{2mm}In this work, we employ the SGD optimizer with a momentum of 0.9, a weight decay of 5 × $10^{-4}$, the $L_{\infty}$-norm for input perturbation, and the $L_{2}$-norm for weight perturbation. We integrate LAP into three commonly used baselines, V-FGSM, R-FGSM, and N-FGSM, respectively. For each of these baselines, we adhere to the configurations provided in their official repository. Regarding our hyperparameters, we set the $\gamma$ as 0.3, and the detailed setting for $\beta$ can be found in Table~\ref{tab:Setting}.

\renewcommand{\arraystretch}{1.0}
\begin{table}[t]
\vspace{-0.6em}
\fontsize{10.0}{12.0}\selectfont
		\caption{Hyperparameter $\beta$ settings for CIFAR-10 and CIFAR-100.}
		\label{tab:Setting}
		\begin{center}
		\begin{tabular}{c|c c c c}
			\toprule
			\toprule
			\multirow{1}{*}{$\beta$} & 8/255 & 12/255 & 16/255 & 32/355  \cr
			\midrule 
			\multirow{1}{*}{V-LAP}& \,\,\,0.03 & 0.058 & \,\,\,0.07 & \,\,\, 0.48\cr
			\multirow{1}{*}{R-LAP}& 0.002 & \,\,\,0.03 & \,\,\,0.05 & \,\,\,\,\,\,0.3 \cr
			\multirow{1}*{N-LAP} & 0.001 & 0.002 & 0.005 & 0.075 \cr
			\bottomrule
			\bottomrule
		\end{tabular}
		\end{center}
\vspace{-2.0em}
\end{table}

\subsection{Performance Evaluation}
\label{section:4_2}
\textbf{CIFAR-10 Results.}\hspace*{2mm}In Table~\ref{tab:CIFAR10}, we report the natural and robust test accuracy of our proposed method alongside the competing baselines. These results are obtained at the final training epoch without the early stopping. From Table~\ref{tab:CIFAR10}, it is evident that LAP demonstrates superior performance across all evaluation cases. More specifically, in the cases where CO does not occur in baselines, our method demonstrates a consistent ability to improve robustness. More importantly, in the cases where baselines are affected by CO, LAP not only effectively prevents its occurrence but also substantially boosts overall performance. It is worth noting that our method can reliably prevent CO even under extreme noise magnitude, underscoring its trustworthy effectiveness.

\textbf{CIFAR-100 Results.}\hspace*{2mm}We also extend our experiments to the CIFAR-100 dataset, wherein the number of categories is increased tenfold and the number of training data per category is reduced tenfold. Notably, CIFAR-100 is more challenging than CIFAR-10, manifested by a greater sensitivity of baseline methods to the occurrence of CO, as shown in Table~\ref{tab:CIFAR100}. Despite the increased challenge, our proposed LAP method consistently demonstrates its effectiveness in mitigating CO and further enhancing adversarial robustness. The above results highlight the reliability and broad applicability of our approach in preventing CO.

\begin{figure*}[t]
\begin{center}
    \begin{subfigure}
    {
            \includegraphics[width=0.66\columnwidth]{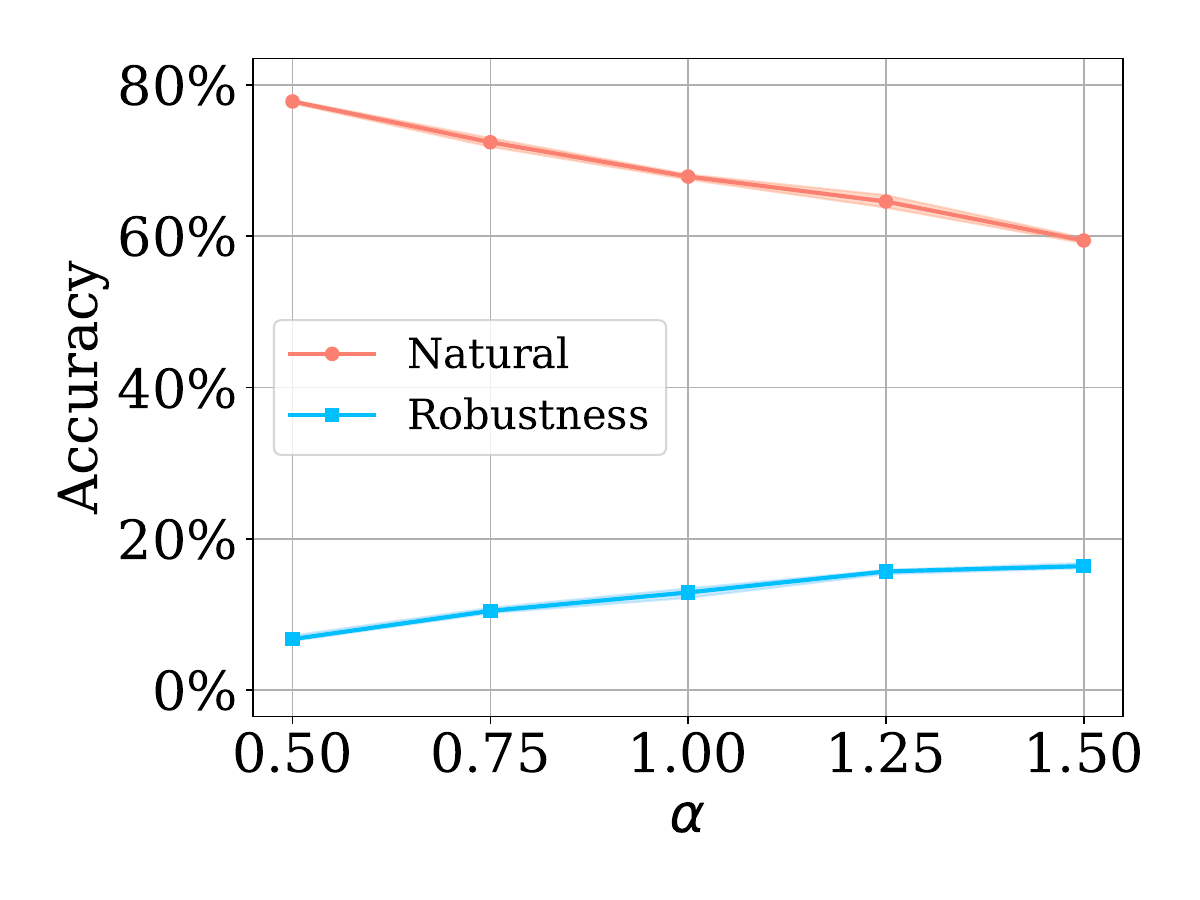}
            \includegraphics[width=0.66\columnwidth]{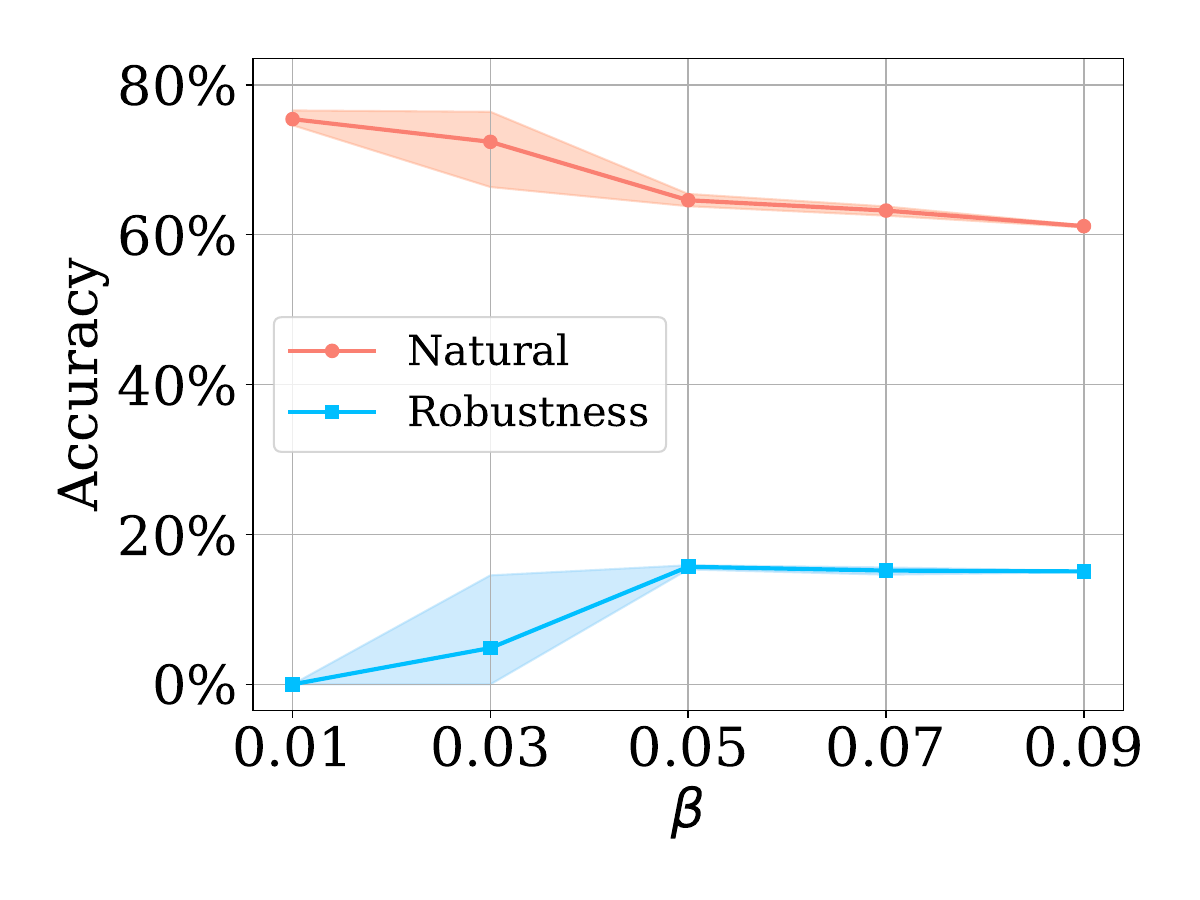}
            \includegraphics[width=0.66\columnwidth]{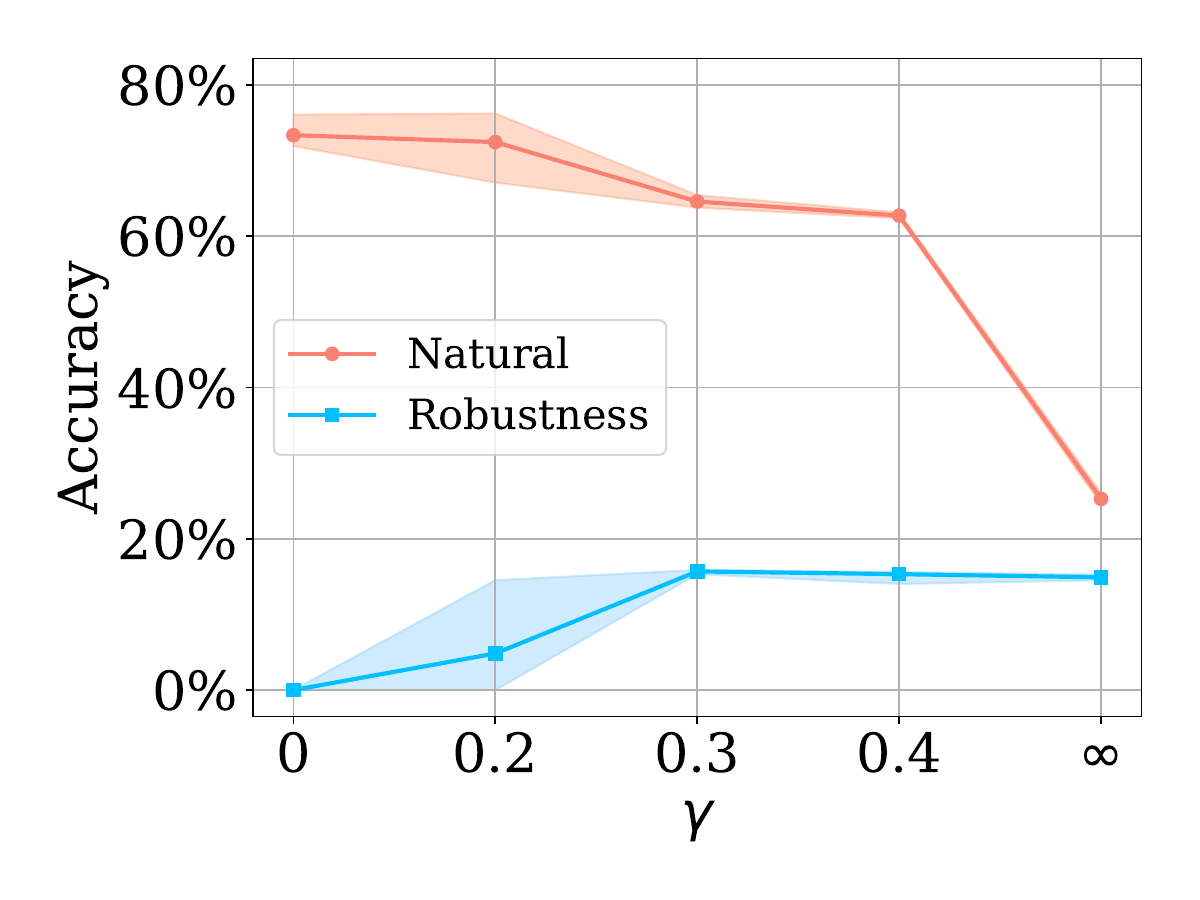}
    }
    \end{subfigure}
\vspace{-1.2em}
\caption{The impact of hyperparameter $\alpha$, $\beta$ and $\gamma$ are shown in the left, middle, and right panels, respectively.}
\label{fig:Role}
\vspace{-1.5em}
\end{center}
\end{figure*}

\begin{table*}[t]
\fontsize{9.3}{12.0}\selectfont
\caption{Comparison of training cost. The results are obtained on a single NVIDIA RTX 4090 GPU and averaged over 30 training epochs.}
\label{tab:Time}
		\begin{center}
		\begin{tabular}{c|c c c c c c | c c}
			\toprule
			\toprule
			\multirow{1}{*}{Method}&FreeAT&Grad Align& ZeroGrad&MultiGrad&V/R/N-FGSM&V/R/N-LAP& PGD-2&PGD-10\cr
			\midrule 
			\multirow{1}*{Training Time (S)} &43.8 &36.1&11.1&21.7&\textbf{11.0}&\cellcolor{LightGreen}11.8&16.4&59.1\cr
			\bottomrule
			\bottomrule
            \end{tabular}
		\end{center}
\vspace{-1.5em}
\end{table*} 

\begin{table}[t]
\fontsize{10.0}{12.0}\selectfont
\vspace{-0.7em}
\caption{Comparison of test accuracy (\%) for LAP with various optimization objectives. The results are averaged over three random seeds and reported with the standard deviation.}
\label{tab:Ablation}
\begin{center}
  \begin{tabular}{l | c c}
    \toprule
    \toprule
    \multirow{1}*{Method} & Natural  & Auto Attack  \\
    \midrule
    \rowcolor{LightGreen}
    \multirow{1}*{LAP}&64.83{\scriptsize$\pm$0.29} & 15.69{\scriptsize$\pm$0.28} \\
    \multirow{1}*{Original AWP}&88.47{\scriptsize$\pm$0.75} & \,\,\,0.00{\scriptsize$\pm$0.00} \\
    \multirow{1}*{Modified AWP}&30.00{\scriptsize$\pm$0.25} & 12.53{\scriptsize$\pm$0.98} \\
    \multirow{1}*{LAP-A}&59.09{\scriptsize$\pm$0.85} & \textbf{15.72}{\scriptsize$\pm$0.01}\\
    \multirow{1}*{LAP-R}&53.87{\scriptsize$\pm$0.14} & 11.22{\scriptsize$\pm$0.10} \\
    \multirow{1}*{LAP-$ L_{\infty}$}&20.38{\scriptsize$\pm$0.38} & 13.67{\scriptsize$\pm$0.35}\\
    \bottomrule
    \bottomrule
  \end{tabular}
\end{center}
\vspace{-1.5em}
\end{table}

\subsection{Ablation Study}
\label{section:4_3}
In this part, we conduct an examination of each component within the R-LAP on CIFAR-10 under 16/255 noise magnitude using PreActResNet-18.

\textbf{Loss Landscape.}\hspace*{2mm}To showcase the effectiveness of our proposed method, we illustrate the loss landscape for both the whole model and individual layers, using the same visualization approach as detailed in Section~\ref{section:3_1}. Compared to the baseline illustrated in Figure~\ref{fig:Landscape}, it clearly demonstrates that LAP leads to a more flattened loss landscape for both individual layers and the whole model, as shown in Figure~\ref{fig:Landscapeour}. This outcome indicates that our proposed method can effectively hinder the generation of \emph{pseudo-robust shortcuts} which typically result in sharp decision boundaries, thereby successfully preventing the occurrence of CO.

\textbf{Optimization Objectives.}\hspace*{2mm}We also explore LAP in conjunction with other optimization objectives. These include the Original AWP as defined in Equation~\ref{eq:AWP}, Modified AWP retaining the accumulated weight perturbation, LAP-A requiring an \textbf{A}dditional backward propagation as outlined in Equation~\ref{eq:LAP-A}, LAP-R plugging the \textbf{R}andom weight perturbation, and LAP-$L_{\infty}$ using \bm{$L_{\infty}$}-norm weight perturbation. To ensure a fair comparison, we conduct a thorough search on the hyperparameter $\beta$ of these methods, and the results are summarized in table~\ref{tab:Ablation}. It is evident that the original AWP is ineffective at mitigating CO due to its inability to disrupt persistent shortcuts. While the modified AWP can mitigate CO, it demonstrates unsatisfactory natural and robust accuracy. This subpar outcome can be attributed to the introduction of redundant adversarial perturbations in the latter layers, which negatively affect the representation learning. Notably, the LAP-family methods, utilizing diverse operations, can effectively obstruct the generation of \emph{pseudo-robust shortcuts}, thereby preventing CO. This comprehensive outcome further verifies our perspective that the model’s dependence on these shortcuts triggers the occurrence of CO. Nevertheless, while LAP-A shows a slight improvement in robustness, its requests additional backward propagation that significantly limits its applicability. Meanwhile, LAP-R and LAP-$L_{\infty}$ fail to achieve a comparable performance to the reported LAP implementation.

\textbf{Hyperparameters Selection.}\hspace*{2mm}We separately explore the effects of $\alpha$, $\beta$, and $\gamma$ on both natural and robust accuracy. When tuning one hyperparameter, the others remain fixed. From Figure~\ref{fig:Role} (left), we can observe that an increase in $\alpha$ leads to improved robust accuracy, but in turn results in a decline in natural accuracy. In light of this trade-off, we follow the original setting and choose not to modify $\alpha$. From the observations in Figure~\ref{fig:Role} (middle), we note that when $\beta$ is set to a small value, the weight perturbation is inadequate to effectively obstruct \emph{pseudo-robust shortcuts} and mitigate CO. However, excessively increasing $\beta$ will cause an over-smoothing model, thereby leading to a decrease in natural accuracy. In Figure~\ref{fig:Role} (right), a similar trend is observed in the adjustment of $\gamma$. When weight perturbation is applied solely to the 1st layer, it fails to effectively hinder the formation of shortcuts. On the other hand, employing uniform weight perturbation across all layers results in a substantial reduction in the natural accuracy.

\subsection{Training Cost Analysis}
\label{section:4_4}

Efficiency is the primary advantage of single-step AT over multi-step AT, offering better scalability to large networks and datasets. Consequently, the computational overhead becomes a crucial factor in assessing the overall performance. In Table~\ref{tab:Time}, we present a comparison of training time consumption among various methods. It is evident that the training cost of the LAP method is comparable to that of the FGSM method, which imposes only a 7\% additional training cost. In contrast, the Grad Align and PGD-10 methods are significantly more time-consuming, being 3 and 5 times slower than our method, respectively. 
\section{Conclusion}
In this paper, we reveal that deep neural networks' dependency on \emph{pseudo-robust shortcuts} for decision-making triggers the occurrence of catastrophic overfitting. More specifically, our investigation demonstrates the distinct transformation occurring in different network layers, with the former layers experiencing earlier and more severe distortion while the latter layers exhibit relative insensitivity. Our study further discovers that this heightened sensitivity can be attributed to the generation of \emph{pseudo-robust shortcuts}, which alone can accurately defend against single-step adversarial attacks but bypass genuine-robust learning, leading to distorted decision boundaries. The model exclusively depends on these shortcuts for decision-making inducing the performance paradox. To this end, we introduce an effective and efficient approach, Layer-Aware Adversarial Weight Perturbation (LAP), which strategically applies adaptive perturbations across different layers to hinder the generation of shortcuts, thereby preventing catastrophic overfitting.

\section*{Impact Statement}
This paper presents work whose goal is to advance the field of adversarial robustness in machine learning. Although single-step adversarial training is the most promising time-efficient method for defending against adversarial examples, it is severely hampered by the catastrophic overfitting problem. In this work, we propose the Layer-Aware Adversarial Weight Perturbation (LAP) method, which aims to effectively and efficiently prevent catastrophic overfitting. Despite LAP being designed to save computing resources, it may still have potential negative impacts on environmental protection (\textit{e.g.}, carbon footprint and global warming). Last and most importantly, while our goal is to develop more secure and robust machine learning for real-world applications, it is crucial to acknowledge that attaining completely safe and trustworthy models is still a distant objective.

\section*{Acknowledgements}
The authors express gratitude to Muyang Li and Suqin Yuan for their helpful feedback. The authors also thank the reviewers and area chair for their valuable comments. Bo Han is supported by the NSFC General Program No. 62376235, Guangdong Basic and Applied Basic Research Foundation Nos. 2022A1515011652 and 2024A1515012399, HKBU Faculty Niche Research Areas No. RC-FNRA-IG/22-23/SCI/04, and HKBU CSD Departmental Incentive Scheme. Hang Su is partially supported by NSFC Projects (Nos. 92248303, 92370124, 62350080). Tongliang Liu is partially supported by the following Australian Research Council projects: FT220100318, DP220102121, LP220100527, LP220200949, and IC190100031.

\bibliography{example_paper}
\bibliographystyle{icml2024}

\newpage
\appendix
\onecolumn

\section{Experiment with WideResNet and Vit Architecture}
\label{appendix:A}

\paragraph{WideResNet-34.}To further validate the effectiveness of LAP, we conduct a performance comparison using WideResNet-34~\cite{zagoruyko2016wide}, which is more complex than PreActResNet-18. In the case of WideResNet-34, we adjust the $\beta$ values for the V/R/N-LAP methods to 0.04, 0.024, and 0.005, respectively, while maintaining other hyperparameters consistent with the original configurations.

\begin{table*}[h]
\caption{Comparison of WideResNet-34 test accuracy (\%) for various methods under 8/255 noise magnitudes on CIFAR-10. The results are averaged over three random seeds and reported with the standard deviation.}
\label{tab:WideResNet}
		\begin{center}
		\begin{tabular}{c|c c c c c c|c}
			\toprule
			\toprule
			Method &V-FGSM& V-LAP & R-FGSM& R-LAP & N-FGSM& N-LAP & PGD-2\cr
			\midrule
			Natural  &86.10{\scriptsize$\pm$1.61} & \cellcolor{LightGreen} 81.92{\scriptsize$\pm$1.14} & 85.21{\scriptsize$\pm$0.78} & \cellcolor{LightGreen} 86.10{\scriptsize$\pm$0.08} & 84.85{\scriptsize$\pm$0.25} & \cellcolor{LightGreen} 84.42{\scriptsize$\pm$0.49} & 88.55{\scriptsize$\pm$0.11} \cr
   			\midrule
                 PGD-50-10  & \,\,\,0.00{\scriptsize$\pm$0.00} & \cellcolor{LightGreen} 44.64{\scriptsize$\pm$0.59} & \,\,\,0.00{\scriptsize$\pm$0.00} & \cellcolor{LightGreen} 46.29{\scriptsize$\pm$0.69} & 49.32{\scriptsize$\pm$0.32} & \cellcolor{LightGreen} \textbf{50.53}{\scriptsize$\pm$0.14} & 46.75{\scriptsize$\pm$0.11}  \cr
			\bottomrule
			\bottomrule
		\end{tabular}
  		\end{center}
\end{table*}

Table~\ref{tab:WideResNet} illustrates that our proposed method, LAP, can consistently prevent CO and achieve a higher level of robustness, comparable to multi-step AT. Moreover, it is worth noting that the complex networks can more significantly demonstrate the efficiency advantages of our method in terms of training time. The results obtained with WideResNet-34 emphasize the applicability of our method in complex network architectures.

\paragraph{Vit-small.}By testing our method on both PreActResNet-18 and WideResNet-34, we have verified its effectiveness in mitigating CO on CNN-based architectures. To further substantiate our perspective and approach, we extend our verification to Transformer-based architectures, specifically Vit-small~\cite{dosovitskiy2020image}. Regarding Vit, the $\beta$ settings are detailed in Table~\ref{tab:VitSetting}, with all other hyperparameters remaining in the original setting.

\renewcommand{\arraystretch}{1.0}
\begin{table}[h]
\fontsize{10.0}{12.0}\selectfont
		\caption{Hyperparameter $\beta$ settings for Vit-small.}
		\label{tab:VitSetting}
		\begin{center}
		\begin{tabular}{c|c c c c}
			\toprule
			\toprule
			\multirow{1}{*}{$\beta$} & 8/255 & 12/255 & 16/255 & 32/355  \cr
			\midrule 
		    \multirow{1}{*}{V-LAP}& 0.003& 0.006& 0.009 & 0.05 \cr
			\multirow{1}{*}{R-LAP}& 0.002& 0.004& 0.006 & 0.04 \cr
			\multirow{1}*{N-LAP} & 0.001& 0.002& 0.003 & 0.03 \cr
			\bottomrule
			\bottomrule
		\end{tabular}
		\end{center}
\end{table}

\begin{table*}[h]
\fontsize{8.6}{12.0}\selectfont
\caption{Comparison of Vit-small test accuracy (\%) for various methods under different noise magnitudes on CIFAR-10. The results are averaged over three random seeds and reported with the standard deviation.}
\label{tab:Vit}
  \begin{center}
  \begin{tabular}{l c c c c c c c c c c c c c }
    \toprule
    \toprule
    \multirow{2}*{Method} & \multicolumn{2}{c}{8/255} && \multicolumn{2}{c}{12/255} && \multicolumn{2}{c}{16/255} && \multicolumn{2}{c}{32/255} \\
    \cmidrule{2-3}
    \cmidrule{5-6}
    \cmidrule{8-9}
    \cmidrule{11-12}
    & Natural  & PGD-50-10  && Natural  & PGD-50-10  && Natural  & PGD-50-10  && Natural  & PGD-50-10 \\
    \midrule
    \multirow{1}*{V-FGSM}& 39.32{\scriptsize$\pm$1.48} & 25.68{\scriptsize$\pm$0.53} && 25.26{\scriptsize$\pm$0.80} & 17.82{\scriptsize$\pm$0.53} && 14.34{\scriptsize$\pm$6.43} & \,\,\,0.00{\scriptsize$\pm$0.00} && 12.68{\scriptsize$\pm$4.28} & 0.00{\scriptsize$\pm$0.00}\\
    \rowcolor{LightGreen}
    \multirow{1}*{V-LAP}& 41.98{\scriptsize$\pm$0.61} & 26.38{\scriptsize$\pm$0.14} && 24.53{\scriptsize$\pm$0.45} & 18.18{\scriptsize$\pm$0.62} && 17.85{\scriptsize$\pm$0.62} & 11.79{\scriptsize$\pm$0.24} && 16.44{\scriptsize$\pm$0.14} & 8.93{\scriptsize$\pm$0.13}\\
    \multirow{1}*{R-FGSM}&45.08{\scriptsize$\pm$0.37} & 26.28{\scriptsize$\pm$0.30} && 28.08{\scriptsize$\pm$0.99} & 18.80{\scriptsize$\pm$0.38} && 23.80{\scriptsize$\pm$1.07} & 14.27{\scriptsize$\pm$0.04} && 13.71{\scriptsize$\pm$2.11} & 0.00{\scriptsize$\pm$0.00}\\
    \rowcolor{LightGreen}
    \multirow{1}*{R-LAP}&46.56{\scriptsize$\pm$0.03} & \textbf{27.06}{\scriptsize$\pm$0.42} && 27.60{\scriptsize$\pm$0.49} & \textbf{19.01}{\scriptsize$\pm$0.24} && 21.72{\scriptsize$\pm$0.14} & \textbf{15.49}{\scriptsize$\pm$0.24} && 17.15{\scriptsize$\pm$0.78} & \textbf{9.04}{\scriptsize$\pm$0.21}\\
    \multirow{1}*{N-FGSM}&37.30{\scriptsize$\pm$1.98} & 24.84{\scriptsize$\pm$0.74} && 24.85{\scriptsize$\pm$0.97} & 17.61{\scriptsize$\pm$0.45} && 20.68{\scriptsize$\pm$0.80} & 13.38{\scriptsize$\pm$1.93} && \,\,\,8.67{\scriptsize$\pm$1.89} & 0.00{\scriptsize$\pm$0.00}\\
    \rowcolor{LightGreen}
    \multirow{1}*{N-LAP}&40.48{\scriptsize$\pm$0.56} & 25.69{\scriptsize$\pm$0.29} && 24.15{\scriptsize$\pm$0.65} & 17.99{\scriptsize$\pm$0.67} && 20.19{\scriptsize$\pm$0.80} & 14.15{\scriptsize$\pm$0.26} && 15.75{\scriptsize$\pm$0.35} & 8.49{\scriptsize$\pm$0.50}\\
    \midrule
    \multirow{1}*{PGD-2}&48.97{\scriptsize$\pm$0.41} & 26.21{\scriptsize$\pm$0.42} && 32.25{\scriptsize$\pm$0.83} & \textbf{19.51}{\scriptsize$\pm$0.28} && 25.42{\scriptsize$\pm$1.00} & \textbf{16.04}{\scriptsize$\pm$0.19} && 18.04{\scriptsize$\pm$3.97} & \textbf{9.67}{\scriptsize$\pm$3.79}\\
    \bottomrule
    \bottomrule
  \end{tabular}
  \end{center}
\end{table*}

It is worth emphasizing that prior research has identified that the CO phenomenon also exists in the Vit model~\cite{shao2022adversarial}, consistent with our observations in Table~\ref{tab:Vit}. Furthermore, the above results underscore two significant differences in the baseline performance between CNN-based and Transformer-based architectures. Firstly, Vit exhibits a lower susceptibility to CO, showing that the V-FGSM does not experience CO when the noise magnitudes are 8 and 12/255, and the R-FGSM can also be effectively trained when the noise magnitude is 16/255. Secondly, the R-FGSM attains the most excellent outcome in baselines, which could be attributed to the larger perturbation introduced by the N-FGSM that disrupts the Transformer-based model learning. Most importantly, Table~\ref{tab:Vit} highlights that our approach can effectively mitigate CO and improve robust accuracy across all levels of noise magnitudes. It is evident both the universality of our perspective and the effectiveness of our approach when applied to Transformer-based architectures.

\section{Settings and Results on Tiny-ImageNet Dataset}
\label{appendix:B}
We also extend our method to a large-sized dataset, Tiny-ImageNet~\cite{netzer2011reading}, to showcase its effectiveness. In the case of Tiny-ImageNet, we set the $\beta$ values for the V/R/N-LAP methods to 0.016, 0.006, and 0.002, while keeping other hyperparameters consistent with their original configurations.

\begin{table*}[h]
\caption{Comparison of Tiny-imagenet test accuracy (\%) for various methods under 8/255 noise magnitudes using PreactResNet-18. The results are averaged over three random seeds and reported with the standard deviation.}
\label{tab:Tiny-imagenet}
		\begin{center}
		\begin{tabular}{c|c c c c c c|c}
			\toprule
			\toprule
			Method &V-FGSM& V-LAP & R-FGSM& R-LAP & N-FGSM& N-LAP & PGD-2\cr
			\midrule
			Natural  &32.70{\scriptsize$\pm$4.55} & \cellcolor{LightGreen} 47.35{\scriptsize$\pm$0.46} & 51.65{\scriptsize$\pm$2.15} & \cellcolor{LightGreen} 50.05{\scriptsize$\pm$0.47} & 48.86{\scriptsize$\pm$0.75} & \cellcolor{LightGreen} 47.82{\scriptsize$\pm$0.24} & 46.58{\scriptsize$\pm$0.45} \cr
   			\midrule
                 PGD-50-10  & \,\,\,0.00{\scriptsize$\pm$0.00} & \cellcolor{LightGreen} 17.64{\scriptsize$\pm$0.61} & \,\,\,0.00{\scriptsize$\pm$0.00} & \cellcolor{LightGreen} 19.03{\scriptsize$\pm$0.18} & 20.58{\scriptsize$\pm$0.49} & \cellcolor{LightGreen} \textbf{20.82}{\scriptsize$\pm$0.20} & 20.42{\scriptsize$\pm$0.39}  \cr
			\bottomrule
			\bottomrule
		\end{tabular}
  		\end{center}
\end{table*}

Table~\ref{tab:Tiny-imagenet} presents the results of LAP applied to the Tiny-ImageNet dataset. These results again substantiate our approach's efficacy in effectively preventing CO and enhancing robust accuracy, establishing it as a dependable solution for large-scale datasets.

\section{Long Training Schedule Results}
\label{appendix:C}
We have further evaluated the performance of our method using the standard multi-step AT schedule~\cite{rice2020overfitting}, which consists of 200 epochs with an initial learning rate of 0.1. The learning rate is reduced by 10 at the 100th and 150th epochs, respectively.

\begin{table*}[h]
\caption{Comparison of long training schedule test accuracy (\%) for various methods under 8/255 noise magnitudes using PreactResNet-18. The results are averaged over three random seeds and reported with the standard deviation.}
\label{tab:Long}
		\begin{center}
		\begin{tabular}{c|c c c c c c|c}
			\toprule
			\toprule
			Method &V-FGSM& V-LAP & R-FGSM& R-LAP & N-FGSM& N-LAP & PGD-2\cr
			\midrule
			Natural  & 87.94{\scriptsize$\pm$0.35} & \cellcolor{LightGreen} 80.11{\scriptsize$\pm$0.18} & 90.89{\scriptsize$\pm$0.76} & \cellcolor{LightGreen} 85.09{\scriptsize$\pm$0.85} & 83.55{\scriptsize$\pm$0.14} & \cellcolor{LightGreen} 83.15{\scriptsize$\pm$0.20} & 86.53{\scriptsize$\pm$0.25}  \cr
			\midrule
                 PGD-50-10  & \,\,\,0.00{\scriptsize$\pm$0.00} & \cellcolor{LightGreen} 31.26{\scriptsize$\pm$0.10} & \,\,\,0.00{\scriptsize$\pm$0.00} & \cellcolor{LightGreen} 36.17{\scriptsize$\pm$0.53} & 36.79{\scriptsize$\pm$0.38} & \cellcolor{LightGreen} \textbf{37.37}{\scriptsize$\pm$0.21} & \textbf{37.99}{\scriptsize$\pm$0.07}  \cr
			\bottomrule
			\bottomrule
		\end{tabular}
  		\end{center}
\end{table*}

Table~\ref{tab:Long} illustrates that our method, LAP, consistently enhances adversarial robustness in the face of another commonly adopted training schedule. This reaffirms the LAP's consistent, reliable, and effective performance in mitigating CO.

\end{document}